\documentclass[journal]{IEEEtran}
\usepackage{graphicx}
\graphicspath{{./}}
\usepackage{setspace}
\usepackage{subfigure}
\begin{document}
\title{A reactive robotized interface for lower limb rehabilitation: clinical results}
\author{Ludovic Saint-Bauzel*, 
				Viviane Pasqui* and Isabelle Monteil**}
\thanks{ e-mail: saintbauzel@isir.fr, pasqui@isir.fr,}
\thanks{imonteil-roch@centresaintemarie.com }
\thanks{*ISIR,UPMC-Paris 6}
\thanks{**Sainte Marie Hospital, Paris}
\thanks{Manuscript received 15 september, 2008; revised december 25, 2008.}
\markboth{Journal of \LaTeX\ Class Files,~Vol.~X, No.~X, January~2009}%
{Shell \MakeLowercase{\textit{et al.}}: Bare Demo of IEEEtran.cls for Journals}
\maketitle
\begin{abstract}
This article presents clinical results from the use of MONIMAD, a reactive robotized interface for lower limb Rehabilitation of patients suffering from cerebellar disease. 
The first problem to be addressed is the postural analysis of sit-to-stand motion.
Experiments with healthy subjects were performed for this purpose.
Analysis of external forces shows that
sit-to-stand transfer can be subdivided into several phases: preacceleration, acceleration, start rising, rising.
Observation of Center of Pressure, ground forces and 
horizontal components force on handles yields rules to
identify the stability of the patient and to adjust the robotic interface motion to the 
human voluntary movement. These rules are used in a fuzzy-based controller implementation. 
The controller is validated on experiments with diseased patients in Bellan Hospital.
\end{abstract}
\maketitle
\section{Introduction}
\begin{figure}[h]
	\begin{center}
		\includegraphics[width=7cm]{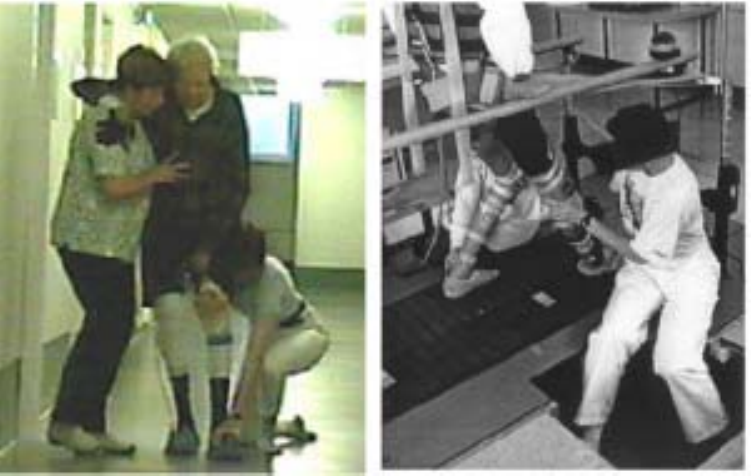}
	\end{center}
	\caption{Classical gait rehabilitation \cite{jezernik03}}
	\label{fig:rehab1}
\end{figure}

Rehabilitation involves the management of disorders that alter the function and the performance of patients.
Rehabilitation is in essence a combination of medication, physical manipulation, therapeutic exercises adapted to technical aids.
In the case of rehabilitation for locomotion, physiotherapists are confronted to an additional problem: management of postural balance. Several persons are then needed to maintain quite at the same time the person in standing up posture and make him/her do therapeutic movements.
This supplementary task is difficult and does not require any medical skills (see Fig. \ref{fig:rehab1}, left ).

In addition, postures needed to apply these exercises to a patient are uncomfortable for medical staff (see Fig. \ref{fig:rehab1}, right).
Consequently, exercises are short in time, a further limitation to the rehabilitation protocols.

Finally, the more time the medical staff spends with a patient, the better the patient is healed but less patients are healed.

Recently, technological aids and robots have been introduced to reduce the number of persons around the patient.

\begin{figure}[h]
	\begin{center}
		\includegraphics[width=4cm]{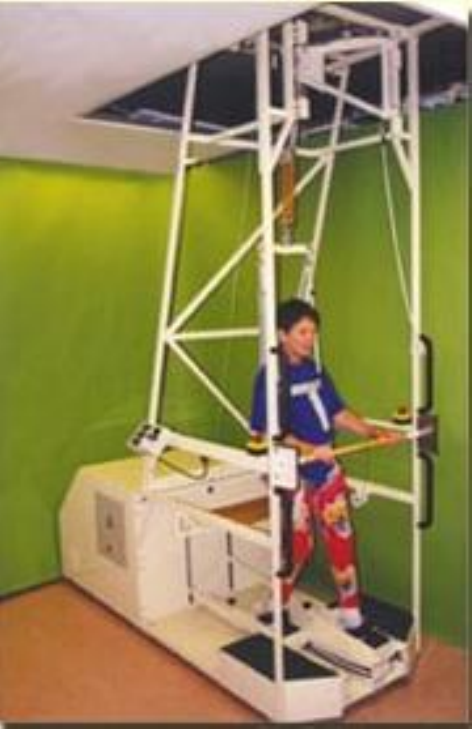}
	\end{center}
	\caption{Rehabilitation with Gait Trainer}
	\label{fig:rehab5}
\end{figure}

Many rehabilitation institutions use electro-mechanical systems such as "Gait Trainer" \cite{hess01}(see Fig. \ref{fig:rehab5}) to solve some of these points. Unfortunately these devices are large that implies they must be installed in a medical institution.  Above all they are only electro-mechanical devices, they have a very limited range of possible protocols, whereas a robotic device from its programming ability could be much more versatile.
\\
Indeed, robotic systems could be an asset and may be used to:
\begin{itemize}
	\item Reduce the load of the medical staff
	\item Make some repetitive basic movements
	\item Assist the patient in therapeutic movements
	\item Guide movements to be as natural as possible
	\item Keep control of therapeutic movements
	\item Develop new rehabilitation protocols
	\item Bring an evaluation thanks to robot sensors acquisition
\end{itemize}
\vspace*{1em}

For these advantages, many rehabilitation robotic devices are proposed.
Obviously a robotized interface for rehabilitation has to be adapted to the kind of pathologies addressed to assist the patient and to make him/her "work" to reduce effects of his/her disease.

In some pathologies (multi-sclerosis, postfall syndrome, etc.), recovery of the locomotion is possible if someone walks with the patient.
These pathologies affect postural balance and consequently lead to many difficulties during both sit-to-stand transfer and walking actions.
In these cases, it is necessary to support the balance.
This support can be a first requirement to involve rehabilitation. In these conditions, it is important to choose a solution that can be used in daily life and that is able to help the patient during gait and sit-to-stand.
Currently, when the locomotion exists but is deficient, the most used technical aid is the zimmer frame.
Such a mechanical system, improved by advanced robotics techniques in order to reinforce walking in safe conditions, could address many diseases or deficiencies.

Concerning rehabilitation of lower limbs, the most common exercises are addressing locomotor system training. These exercises are used to train upright posture and walking movement and paraplegia patients are often the aim of this therapy.

The robotized solutions for those exercises consist in one hand of a body harness supporting the patient's weight and on the other hand of a robotized interface in contact with lower limbs to make him/her walk.

A first kind of such solutions is based on an exoskeleton structure, existing solutions are Lokomat \cite{reiner06a}, and also AutoAmbulator \cite{kathryn} or PAM/POGO \cite{galvez05}. They are mechanically designed to follow many parts of the body. They bring some asset in guiding. The walking is trained but exoskeleton solutions need too much power to be embedded so that the patient is walking on a threadmill and his motion is guided by the robot. This solution is safe but can only be used in a clinical environment.
For the same reason, a device like HapticWalker \cite{schmidt05b} that is totally different in its design is not suitable. It can only move the feet of the patient. Its mechanical design is based on an analysis of operational space of the feet considered as end effectors that the robot must be able to follow. A weight support is included in the system and it is able to propose some motions of daily life like climbing stairs, walking... 
Those solutions are real clinical aid but due to their great size, they are not suitable for a daily home training. So they are more used for patient that need to recover basic movements. Their lack of mobility does not permit to make daily home reinforcement rehabilitation exercises.

Adapted robotized interfaces like KineAssist \cite{peshkin05} or WHERE \cite{lee02} can help patients that need to walk and to have a  weighty support. However, when the patient is still strong enough to support his/her body, it is not suitable to use a harness, that can lead to a loose of muscular strength.

\begin{figure}
\subfigure[FES Supported sit-to-stand rehabilitation robot]{
 \label{fig:kam}
 \centering
 \includegraphics[width=0.22\textwidth,bb=0 0 430 446]{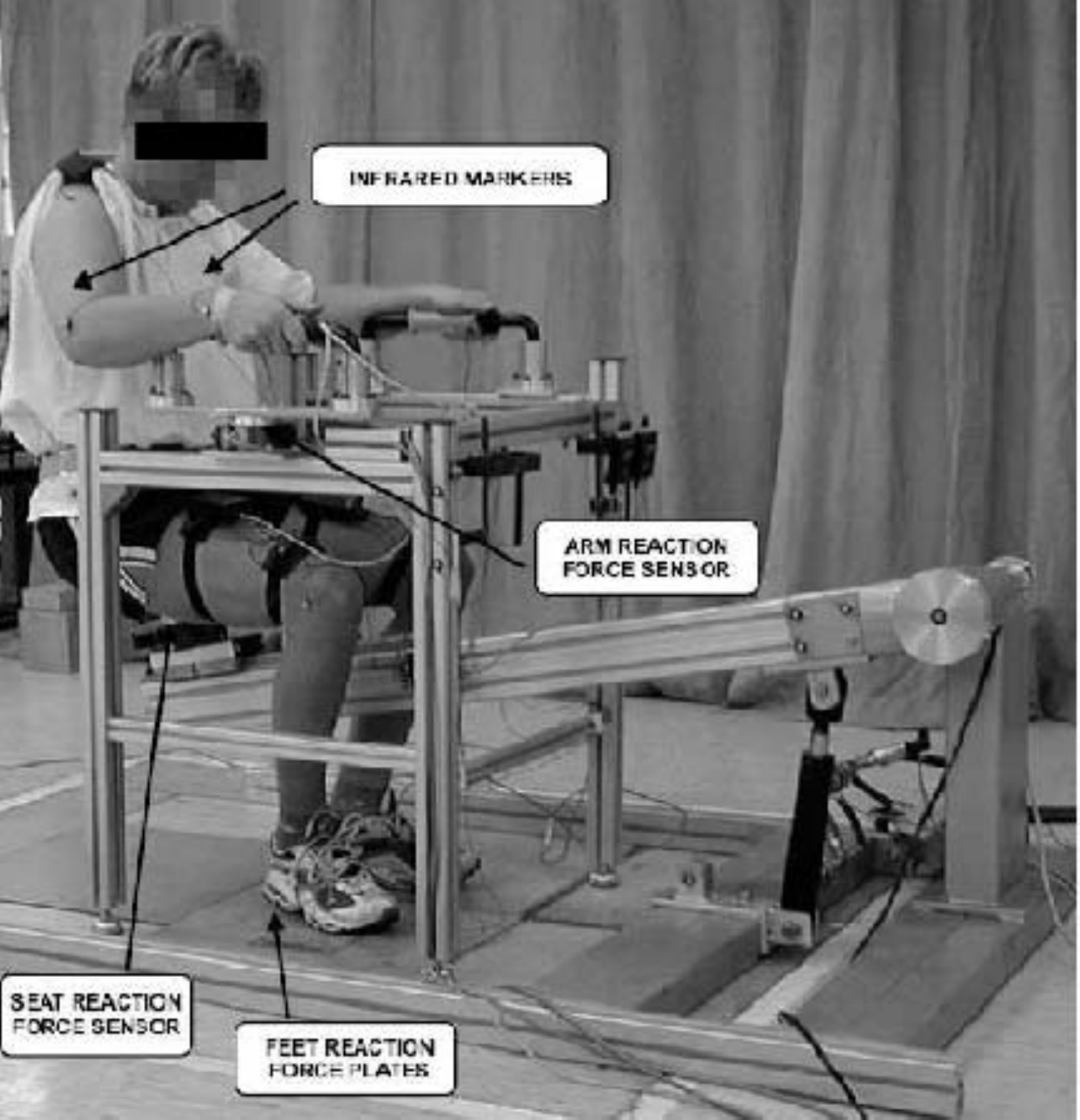}}
%
\subfigure[Standing Assistance System]{
 \label{fig:chugo}
 \centering
 \includegraphics[width=0.22\textwidth]{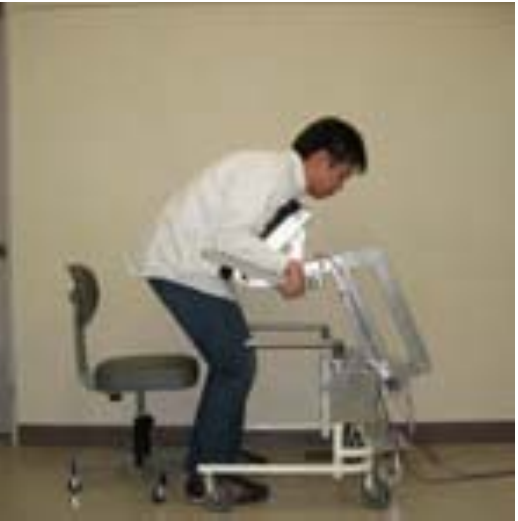}}
\caption{Sit-to-stand robotized solutions}
\label{fig:stsrob}
\end{figure}

If we address sit-to-stand motion, Kamnick and Bajd \cite{kamnik04} propose a rehabilitation robotized solution, that is composed of a robotized chair and a force sensor instrumented handrail (Fig. \ref{fig:kam}). This solution is not mobile so it can only be used in clinical environment.
The  ``Standing Assistant System'' proposed by Chugo \cite{chugo07}, is mobile so it may be a solution for daily life. However the current prototype is designed on a free wheel mobile platform (see Fig. \ref{fig:chugo}) so this prototype is limited to problems with sit-to-stand motion.  It supposes that disbalance during gait could be resolved by a zimmer frame.

\begin{figure}[h]
	\begin{center}
		\includegraphics[width=6cm]{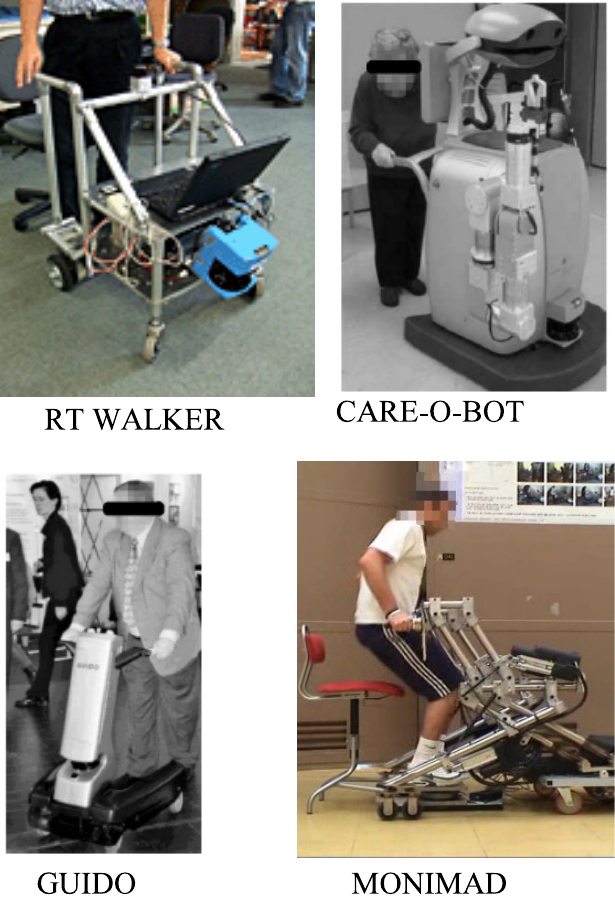}
	\end{center}
	\caption{Robotic walker aids}
	\label{fig:walkers}
\end{figure}

The kind of suitable robotics solution designs able to bring an asset to life of patient that we address in this paper are coming from research that are dedicated to rehabilitate and to assist gait for elderly as: Care-O-Bot \cite{graf01a}, Guido \cite{Rodriguez05}, Walker RT \cite{hirata04} or MONIMAD \cite{mederic05} that are presented in Fig. \ref{fig:walkers}.
The last robot (MONIMAD) is designed and used in the experiments presented here.

The MONIMAD prototype was initially designed to support elderly patients affected by post-fall syndrome \cite{mederic03}. To fit these needs, the main idea is to get inspired by the functionalities of a zimmer frame, improved by contribution of advanced robotics techniques.

\begin{figure}[h]
	\begin{center}
		\includegraphics[width=5cm]{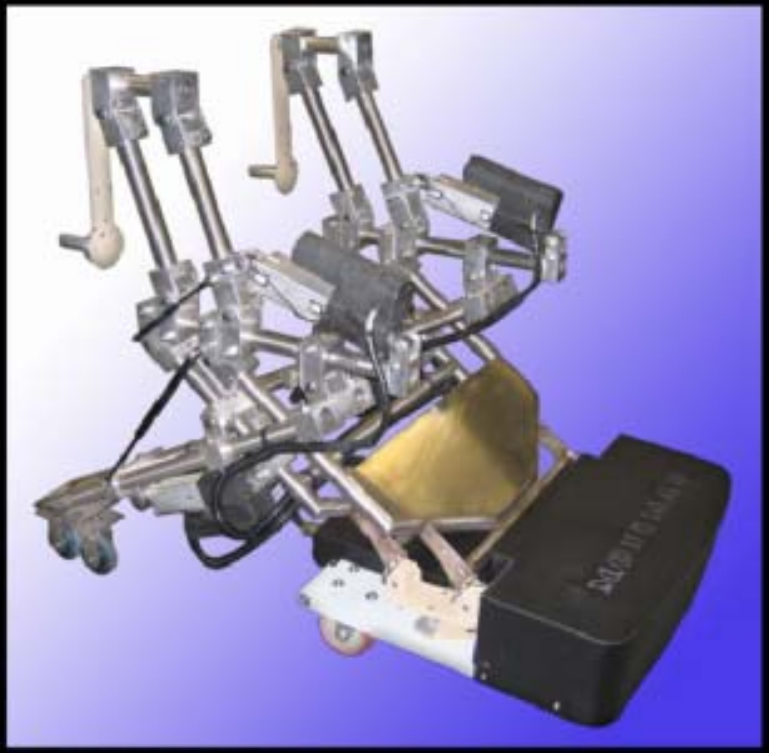}
	\end{center}
	\caption{The MONIMAD Prototype}
	\label{fig:neuneu}
\end{figure}

The robotic device presented in this paper is an active mobile base platform with actuated articulated arms and driven by a whole sensors based control. The control, detailed in this paper, is a reactive control able to identify voluntary movements. Our goal is that the person feels helped by the system rather than driven or guided by a machine.\\
A particularity of this work is that it is centered on helping people. The patient is not considered in the control as a master nor a slave of the robot.
Patients do exercises with the robot in a way that the support of the machine feels transparent.\\
This aim is achieved by the use of a fuzzy-logic based control that works from an immediate and natural handling of the robot, not a control based on a box with buttons or particular gesture to control the device.
Furthermore, assistance must begin from the sit gesture, with as few preparation as possible to use the robotized interface. \\
The MONIMAD prototype (see Fig. \ref {fig:neuneu}) is evaluated in a rehabilitation hospital specialised in the case of multi-sclerosis diseased patients who are often affected by cerebellar ataxia, a disease that leads to trouble in balance during sit-to-stand and walking gestures.

The aim of this paper is to present experiments with MONIMAD used by patient to stand-up, and to study the learning rate of the device.
Section \ref{sec:MatMeth} explains the mechanical structure of the MONIMAD prototype and the implemented control that brings reactivity to the robot.
In section \ref{sec:ResDis}, we describe the experimental protocol worked out by the medical staff.
Then, section \ref{sec:discuss} is dedicated to the discussion on the pros and cons of both our method and our experiments.

\section{Material and Methods}
\label{sec:MatMeth}
The aim of the MONIMAD prototype is to help people without human assistance. This work is driven by the Physical Human-Robot Interaction in mechanical design and in control design. The design of this solution can be divided in two main parts:
\begin{itemize}
\item the assistive robot with its mechanical characteristics,
\item the control that makes the robot actions intuitive and safe for patients.
\end{itemize}
A particularity of this work is that it is centered on helping people. The role of the patient in the control is neither to be a master nor a slave of the robot but a few of each.

\subsection{Mechanical design}
\begin{figure}[ht]
	\begin{center}
		\includegraphics[width=8.8cm]{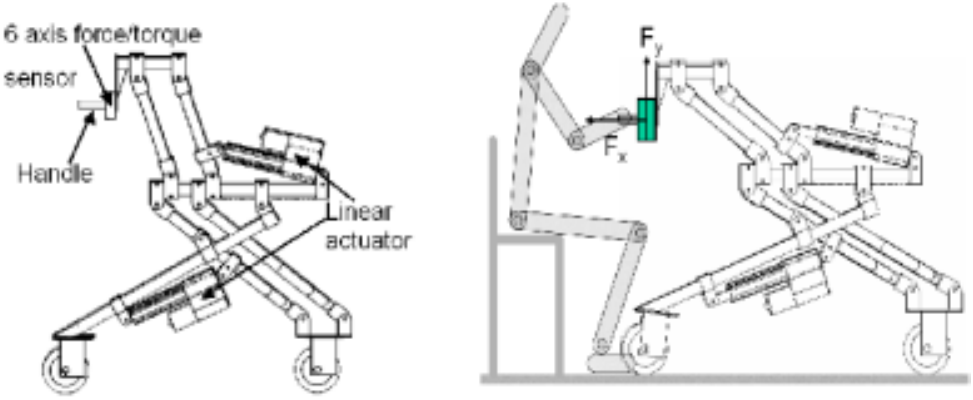}
	\end{center}
	\caption{Description of the robotized interface}
	\label{fig:rehab4}
\end{figure}

The detailed mechanical design method is described in \cite{mederic04a}. The main idea of the design is to place natural actions addressed above as the main requirement.
The designed robotic system is basically a two degrees of freedom (dof) arm mechanism mounted on an active mobile
platform. Its kinematics is described in Fig. \ref{fig:rehab4}.

For the sit-to-stand transfer, handles must first pull slowly the patient to an antepulsion configuration. Then, the handles go from its down to its up position, used for walking. Obviously, the handles must remain horizontal during the whole transition. This is obtained by a serial combination of two 1 dof closed loop mechanisms. The upper part of the mechanism is constituted by two simple parallelograms: the arms and the lower part
is equivalent to a Scott-Russel mechanism \cite{chang05}.

The arms are independent in order to restore lateral balance when the
user begins to lose it, this functionality is not presented in this paper. The wheelbase length is variable: it is longer to increase stability during the sit-to-stand transfer and shorter during walking for eased ambulation. In addition, handles are equipped with six components forces
sensors that are used to make the whole mechanism transparent to the user (i.e. for Physical Human-Robot Interaction).

Measurements are done on sit-to-stand transfer.
The chosen force range are based on the measured forces of the support platform that helps people to stand-up. These forces are lower than the weight of the patient. The robot is not designed to replace the patient motion but to bring some force to support him/her during his/her own motion.

\subsection{Control design}
\begin{figure}[ht]
	\centering
		\includegraphics[width=6cm]{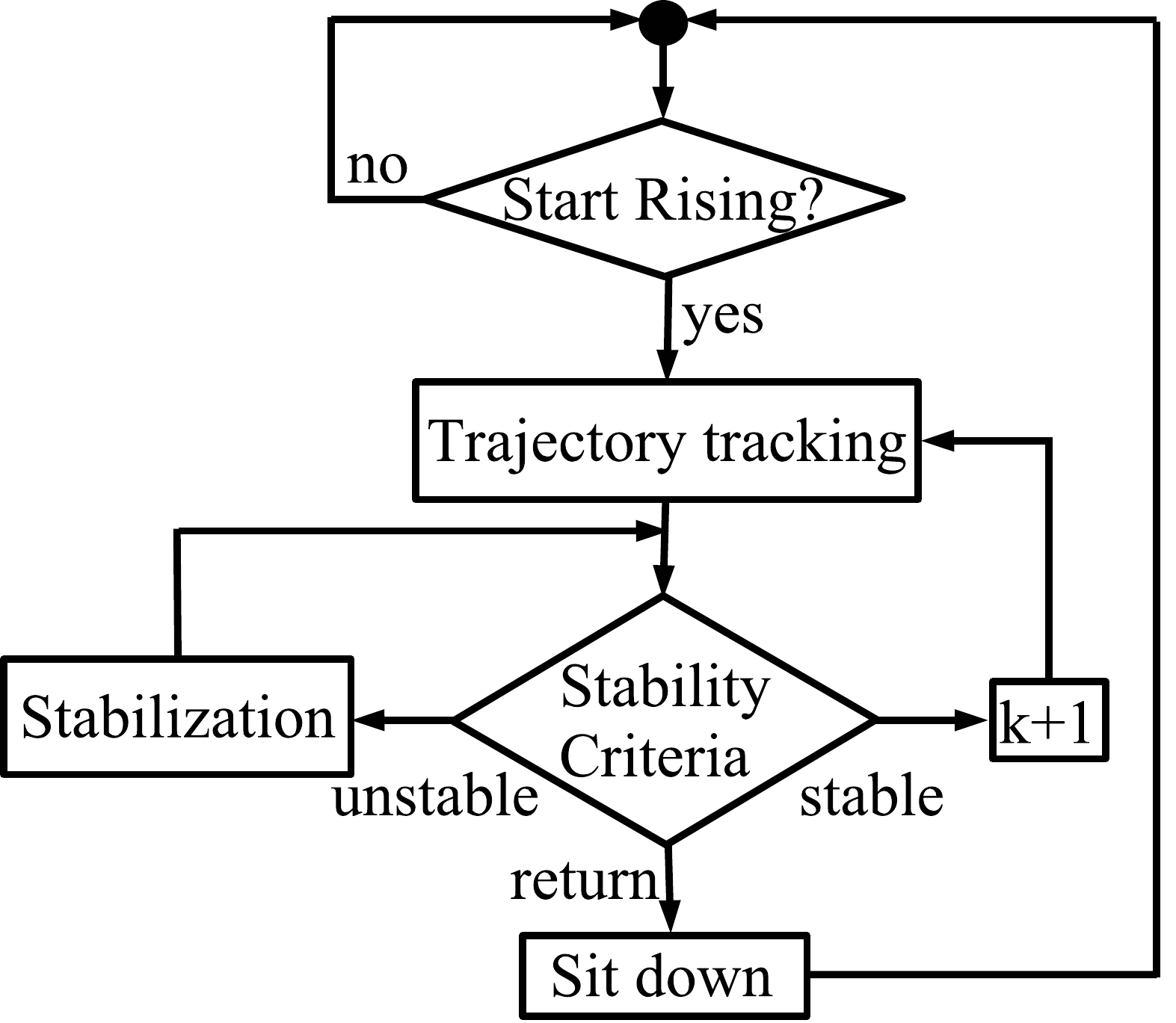}
	\caption{A sit-to-stand scenario schema}
	\label{fig:sts}
\end{figure}

In this section, we explain how an adapted control can give interactive ability to this robotic interface.
By interactivity, we mean the capacity to interpret the postural movements
detected by the sensors to trigger the movement or
to maintain postural balance.\\
In a normal sit-to-stand scene, the patient puts his weight on the robot handles, rises up from the chair and walks.
But many others cases can appear in the scene such as: the patient cannot rise from the chair and wants to seat back or when he is nearly standing up, he loses balance, etc. \\
How do we detect these abnormal cases and what does the robot do?
The different ways to detect these abnormal cases and the corresponding robot reactions are presented in Fig. \ref{fig:sts}.
\\
The detection of these abnormalities is based on human postural analysis, thus the different reactions of the robot are control laws and the overall schema is managed by a fuzzy supervisor.

\subsection{Abnormality detection}
\label{sec:abdetec}
To observe the postural state, experimental dynamical analysis of the stand-up gesture has been performed in our laboratory \cite{pasqui07}.
To record postural data, subjects were instrumented with goniometers placed on the leg articulations (hip, knee, ankle) and accelerometers placed on the breast. We also used an instrumented handle equipped with a 6 axis force sensor and a localization sensor (MiniBird). In addition, the subject's feet were placed on a 6 axis force sensor.

\begin{figure}[ht]
 \centering
 \includegraphics[width=0.30\textwidth,bb=0 0 337 398]{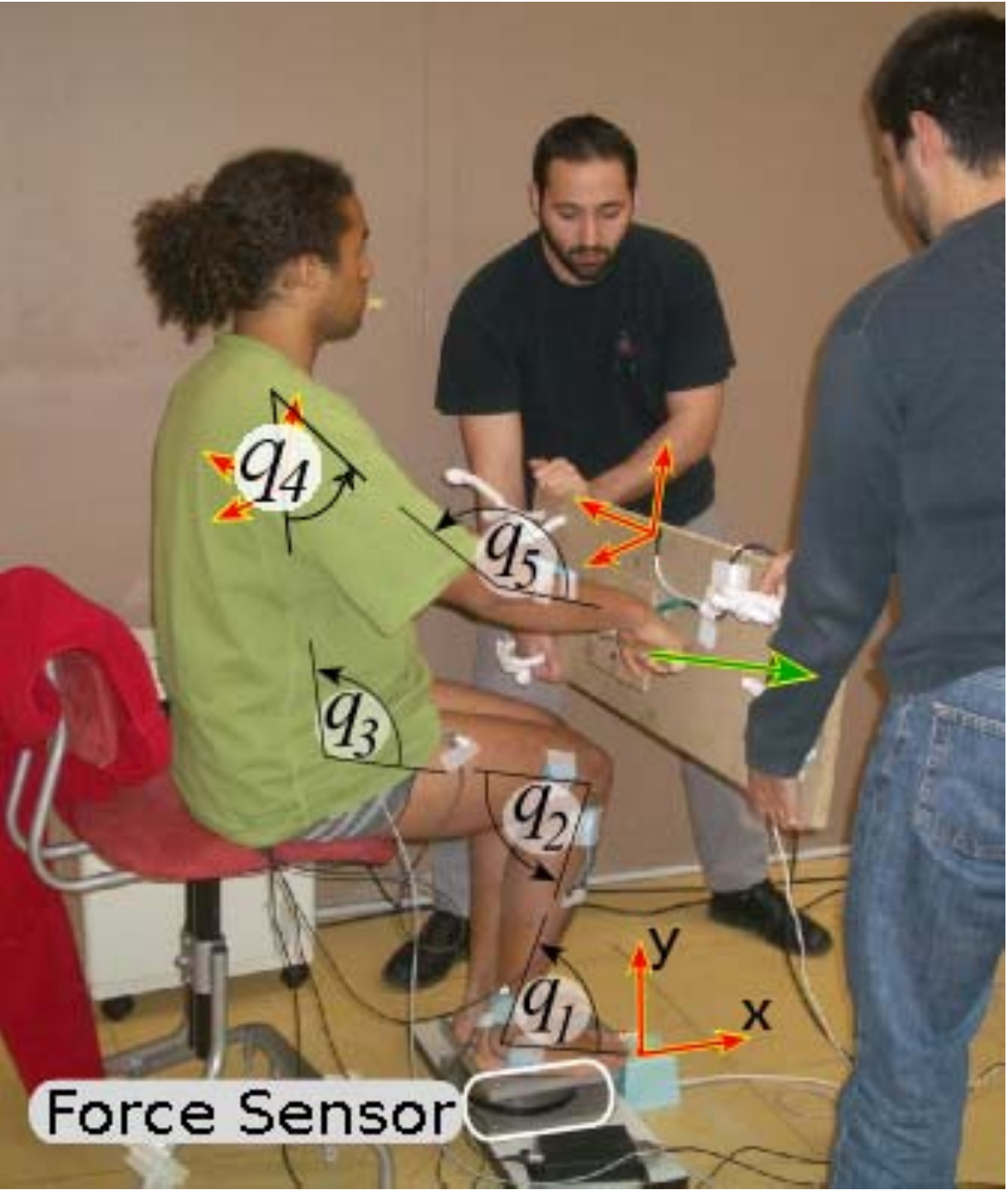}
 \caption{First phase of recording}
 \label{fig:FirstExperimentalRecords}
\end{figure}

Figure \ref{fig:FirstExperimentalRecords} shows the experimental set-up for these recording experiments.

Subjects are 10 healthy people of 25 years in average, weighting 70 kg. They are equipped and placed on a chair. They are asked to hold the instrumented handle and to try to stand-up. Subjects are invited to realise two gestures:
\begin{itemize}
 \item 10 natural speed sit-to-stand,
 \item 10 high speed sit-to-stand (as fast as they can without loosing contact with the ground).
\end{itemize}

In order not to exhaust the subjects, they are advised to make a long time pause between each movement.

\begin{figure}[htbp]
\begin{center}
\subfigure[3 phases of sit-to-stand movement]{%
	\label{fig:phases}
	\includegraphics[width=0.15\textwidth]{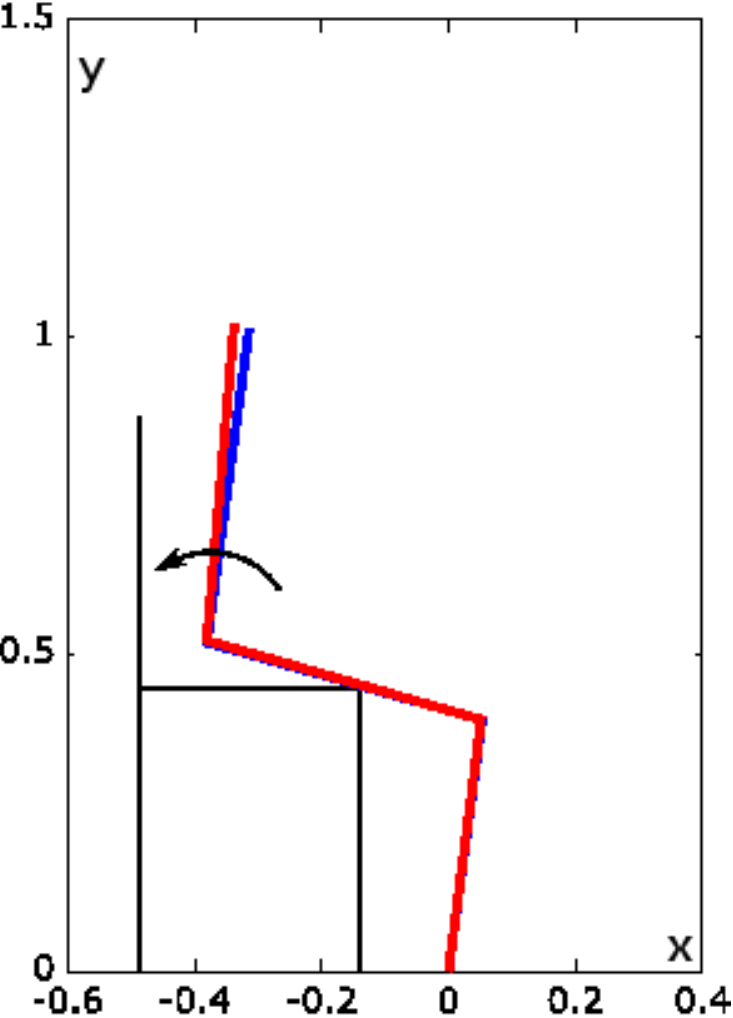}
	\includegraphics[width=0.15\textwidth]{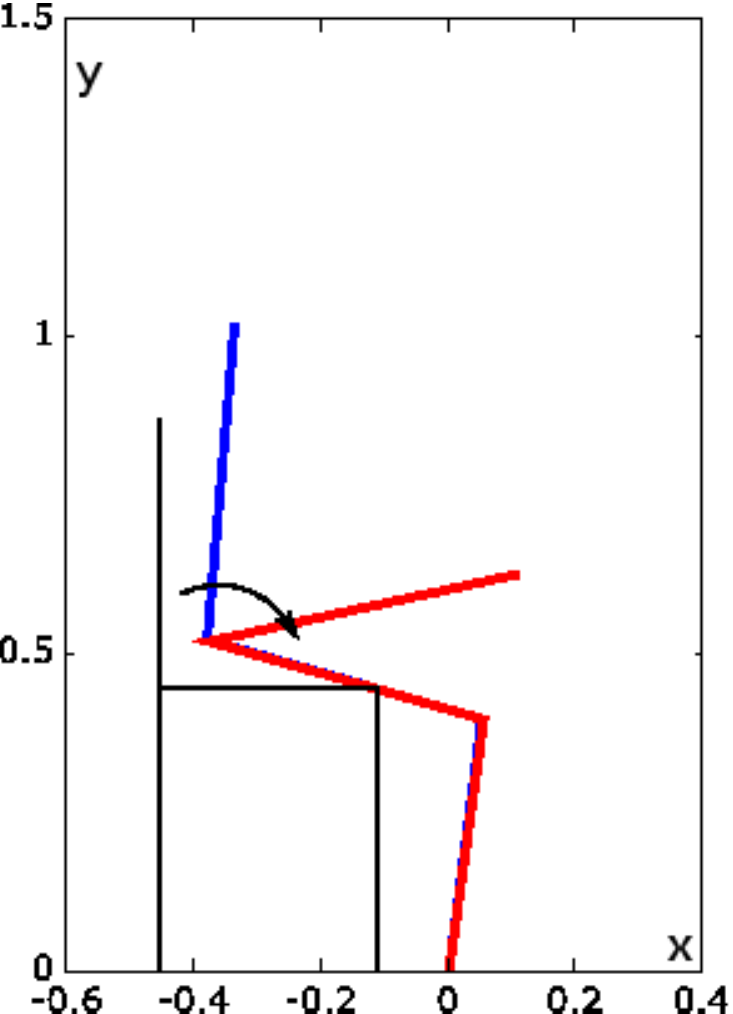}
	\includegraphics[width=0.15\textwidth]{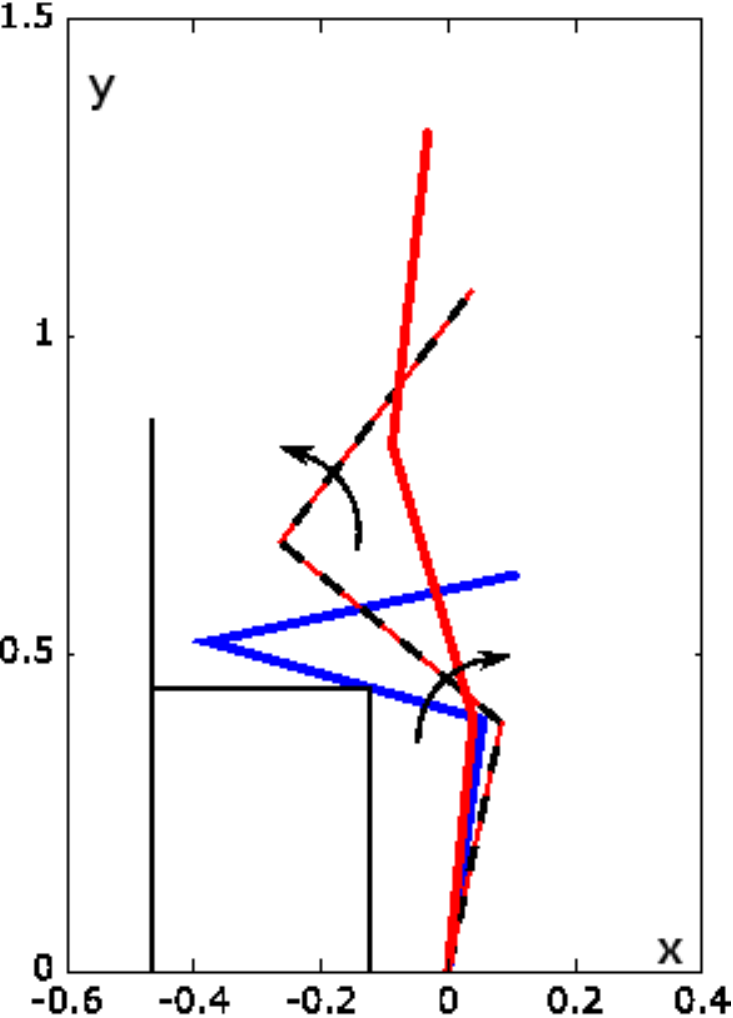}}
\subfigure[Forces Recorded and their derivatives]{%
 \label{fig:recStab}
\includegraphics[width=0.45\textwidth]{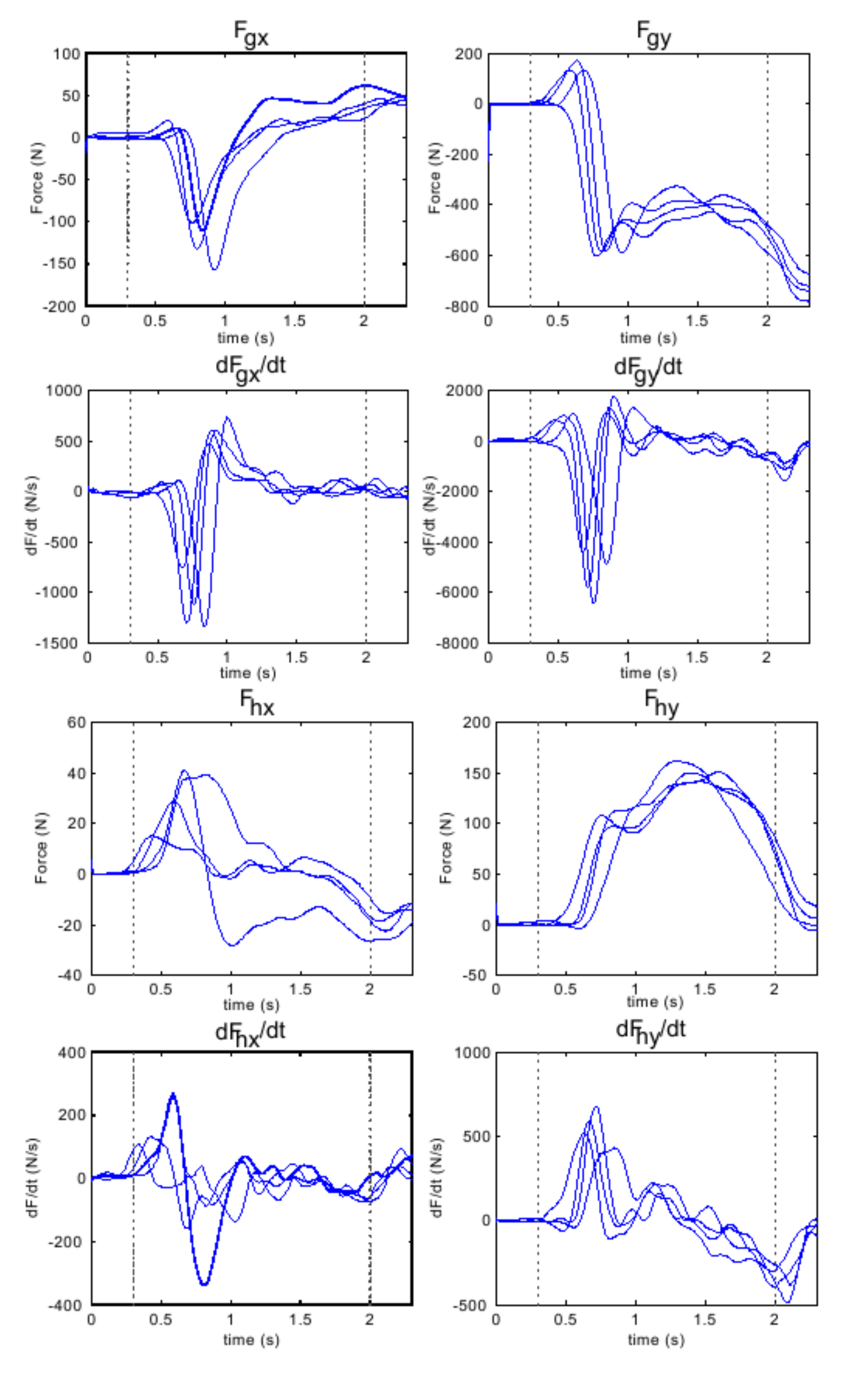}}
\end{center}
	\caption{Different sit-to-stand phases analysis}
\label{fig:stsana}
\end{figure}

Results in Fig. \ref{fig:stsana} show different phases of chair rising, that match with physiological literature \cite{aissaoui99}.

%

Three main sit-to-stand phases are represented in Fig. \ref{fig:phases}. These phases are: pre-acceleration, acceleration and rising.
Each phase depends on interaction forces between the subject and the handles: $\vec{F}_h=(F_{hx},F_{hy})$, the subject and the ground:
$\vec{F}_g=(F_{gx},F_{gy})$ and their time variations.
The Center of Pressure (CoP) which position may be used as a stability criterion \cite{sardain04b} is computed from the reaction force.\\
Observation of the CoP position and direction of the force $\vec{F}_h$ yields
simple rules to identify instability cases or desired movement to trigger (i.e. beginning of the sit-to-stand).

\begin{figure}[ht]
	\begin{center}
        \includegraphics[width=6cm]{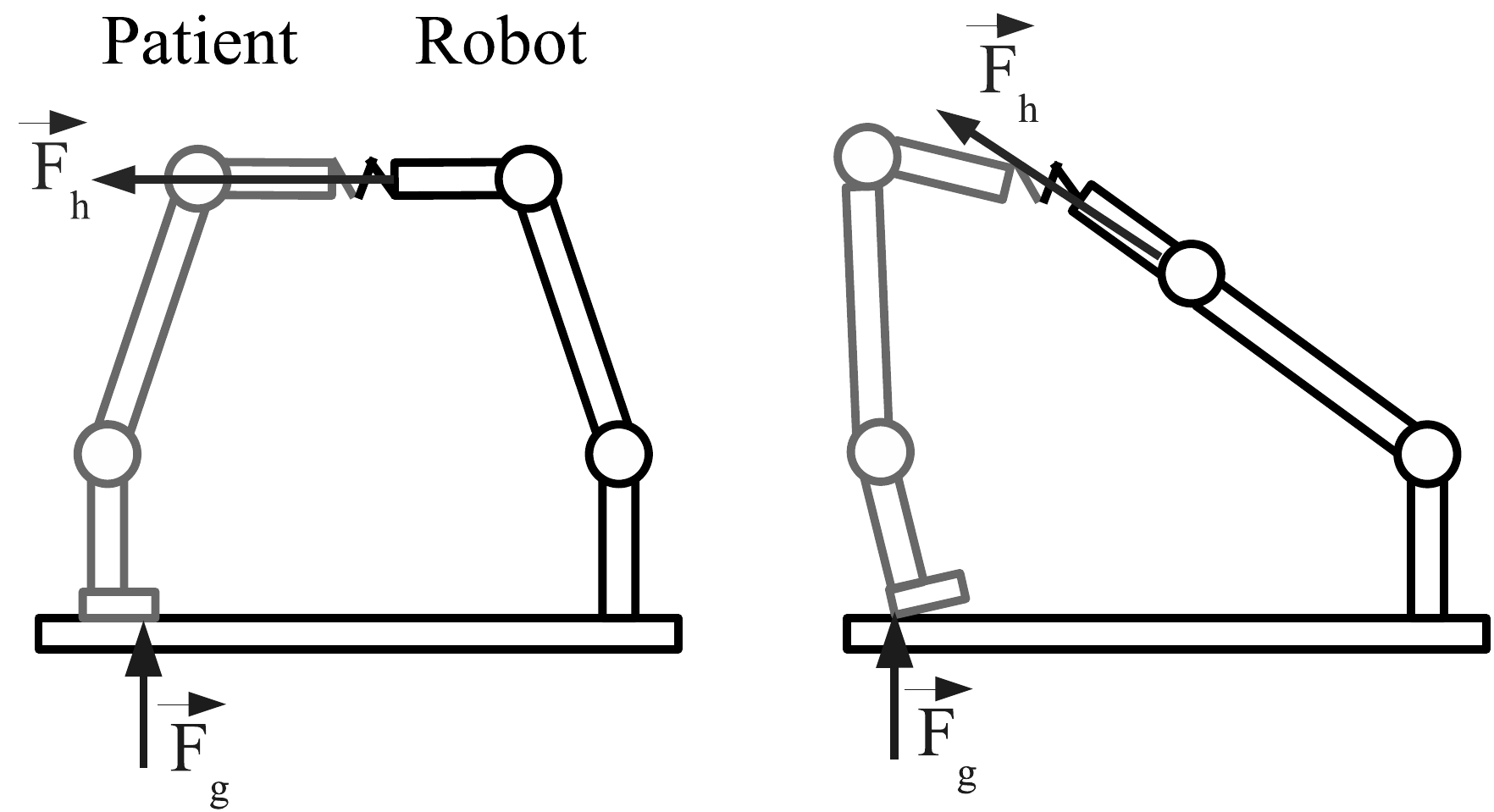}
\end{center}
	\caption{Interaction between patient and robotic interface}
	\label{fig:interaction}
\end{figure}

Detection of unstable posture is illustrated in Fig. \ref{fig:interaction},
where both patient and robot are modelled by a 3 links model each.
The difference between these 2 models lies in the interaction with the ground.
We assume that the robotic interface cannot loose
contact with the ground while the subject could if he or she is unstable.

If a subject, under perturbations, is about to loose balance, he or she quickly shifts the load within the foot support area in the opposite direction with respect to the fall direction (Fig.\ref{fig:interaction}. left). If the perturbation is too high or if the fall is impending, the CoP will rapidly move in the direction of fall until it reaches the limit of the sustentation area (Fig. \ref{fig:interaction}, right).

This supervisor is defined using a set of data obtained with human-human interaction. So from these data the supervisor is tuned for human-robot interaction.

\subsection{Robot reactions as control laws}
\label{sec:laws}
\begin{figure}[ht]
	\centering
		\includegraphics[width=5cm]{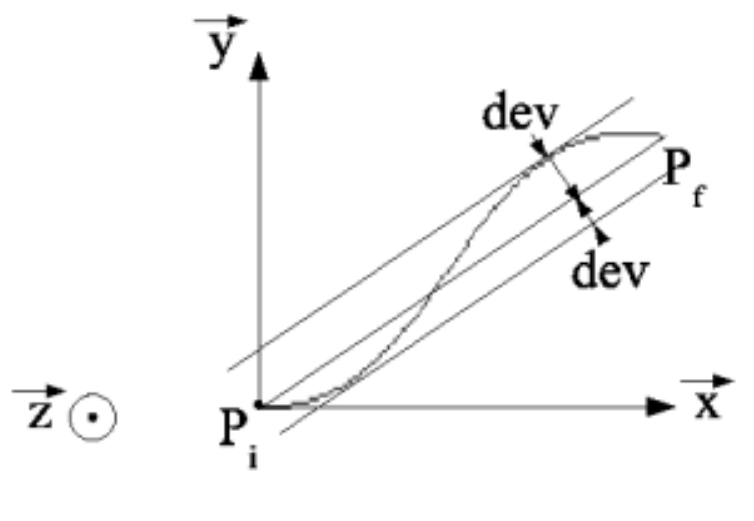}
		\caption{The handles trajectory}
	\label{fig:path2}
\end{figure}

The control is based on different states of the patient that involve different states of the robot. These states are in a higher level than states used in state based control. So one can call these states ``Control Modes''.
Many dangerous situations remains during
the sit-to-stand movement.
It is clear that all of the dangerous situations could not be covered by the
proposed system. For example, in case of fainting the system cannot act. \\
This paper have investigated the lake of balance. In many cases,
these discomfort situations are transients. Then, for these situations the action of the system
on the human posture is sufficient to secure the movement, as a nurse helps with his/her hands.\\
When the nurse's hands "feel" an instability, they first stop the movement and after push/pull the person
in a stable posture.\\
To lift softly a person, the handle movement has to be as close as possible to those induced
to the assistive movement produced by a nurse.
The handles trajectory has to be similar to the general curve presented in Fig.\ref{fig:path2},
where:
\begin{itemize}
	\item $P_i$ and $P_f$ are initial and final handle positions
	\item $dev$ is a physiological parameter characterising the impairment, fitted to the user.
\end{itemize}
The method presented in \cite{pasqui06} consists in decomposing the trajectory characteristics into a physiological part and a mechanical part. \\
Minimum jerk is a physiological constraint for smoothness and infers only the trajectory quality.
Thus, curvilinear abscissa is used to describe the law of motion satisfying minimum jerk.\\
Geometrical path describing hand or some other end effector trajectories are not time dependent and may be expressed in term of Euclidean coordinates.\\
At last, the trajectory is a time function of Cartesian coordinates generating a smooth movement.\\
The control modes implemented in this paper are:
\begin{enumerate}
	\item \textbf{Normal}: The assistive device handles guide the patient to rise from a chair or to sit down,
following trajectories that are based on parameters reflecting personal strategies \cite{mederic06}.
	\item \textbf{Admittance}: To define his or her personal trajectory, the patient must choose the high and the low positions of the handles. To choose these positions, a nurse helps the user to stand-up and the assistive device is in a transparent mode (i.e. the force applied to patient are controlled to be equal to zero).
	\item \textbf{Stabilization}: The handles stop moving and pull/push the hand in the opposite direction w.r.t. the started fall. Then, the tracking trajectory is modified to stabilize the patient as for example Fig.\ref{fig:path3}
\begin{figure}[h]
	\centering
		\includegraphics[width=7cm]{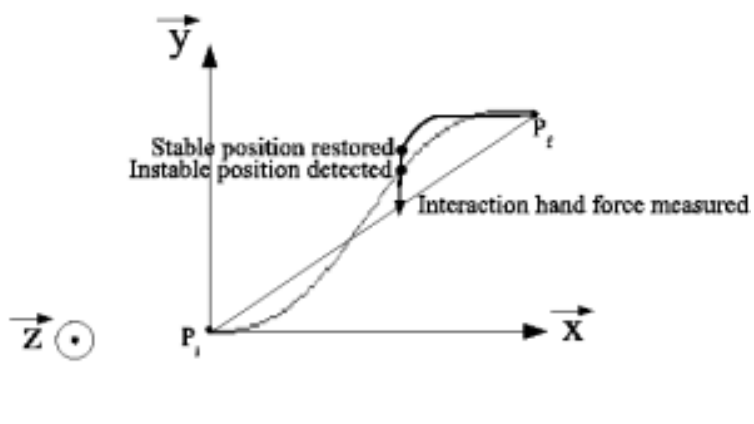}
				\caption{The handles trajectory for keeping the stability}
	\label{fig:path3}
\end{figure}

	\item \textbf{Return}: The interface returns to the initial position following a specific trajectory defined in \cite{mederic06}.
\end{enumerate}
Those control modes are decided to fit with sit-to-stand motion but one could extend the approach to other modes in different rehabilitation contexts.

The robot is programmed to apply a process in 2 phases.
First, automatic positioning of the handles is obtained with an admittance control of one handle sensor.
Then the  proposed exercise is a sit-to-stand motion.
To control that motion, patients are installed on a chair with feet on a 6-axis force sensor. Their hands are holding handles of the assistive device. They are asked to stand-up whenever they want, as far as the supervisor is able to identify initiation of sit-to-stand motion.

A normal sit-to-stand scenario involves control modes as follows. Initially the \textbf{admittance} mode is activated, it lets the patient set the initial position of the handle in order to be in good position for sit-to-stand motion. After that, the \textbf{Normal} mode is triggered by detection of the pre-acceleration phase.
So in that mode, the handlers movement begins to guide the patient from sit to stand posture.
During this sit-to-stand motion, the \textbf{Return} mode can be activated when an aborting movement of the patient is identified. In that case, the robot
returns to the initial position. \\
If postural instability is detected, the \textbf{Stabilization} mode is then activated, the vertical motion of the device is stopped, and a new desired position is computed that guarantees patient stability.\\

All these control modes are designed with fuzzy logic blocks that identify the postural state of the
patient, and put the robot into the corresponding control mode.

\subsection{Fuzzy supervisor}
\label{sec:fuzzy}
\begin{figure}[ht]
\begin{center}
		\includegraphics[width=0.49\textwidth]{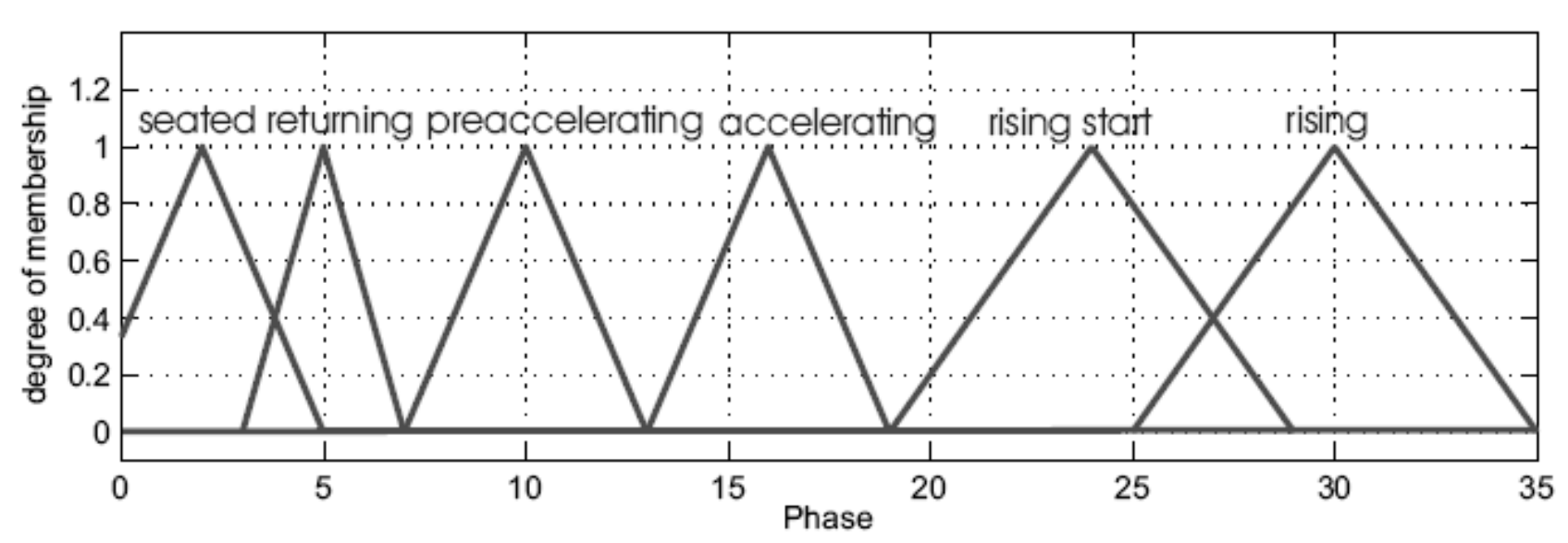}
	\end{center}
	\caption{Membership functions for output 1}
	\label{fig:set1}
\end{figure}

A fuzzy controller is a good way to design an interactive device \cite{sweeney06},\cite{hussein02}. Here, we have extended the role of the fuzzy supervisor from the detection of voluntary movements to the detection of  instability.\\
From the set of experimental data, fuzzy logic sets are tuned to have a representative definition of supervisor.
The fuzzy supervision has to fulfill two tasks, that define two outputs:
\begin{itemize}
	\item \textbf{output 1}: recognition of the current phase, resulting in the choice of control modes 1, 2, 4
	\item \textbf{output 2}: determination of the proper reaction to ensure stability of the subject, and determine amount of use of control mode 3.
\end{itemize}
The fuzzy sets defined for the output 1 are shown in Fig. \ref{fig:set1}

The detection of the phases of the sit-to-stand is obtained analyzing the value of the $\vec{F}_h, \vec{F}_g$ forces shown in Fig. \ref{fig:stsana}, their time variation and computed CoP.

A fuzzy-controller able to represent sit-to-stand transfer is set-up from force information Fig. \ref{fig:recStab} obtained at the handle, force information coming from the ground interaction and specially computation of the CoP.

\begin{figure}[htbp]
\begin{center}
	\includegraphics[width=0.49\textwidth]{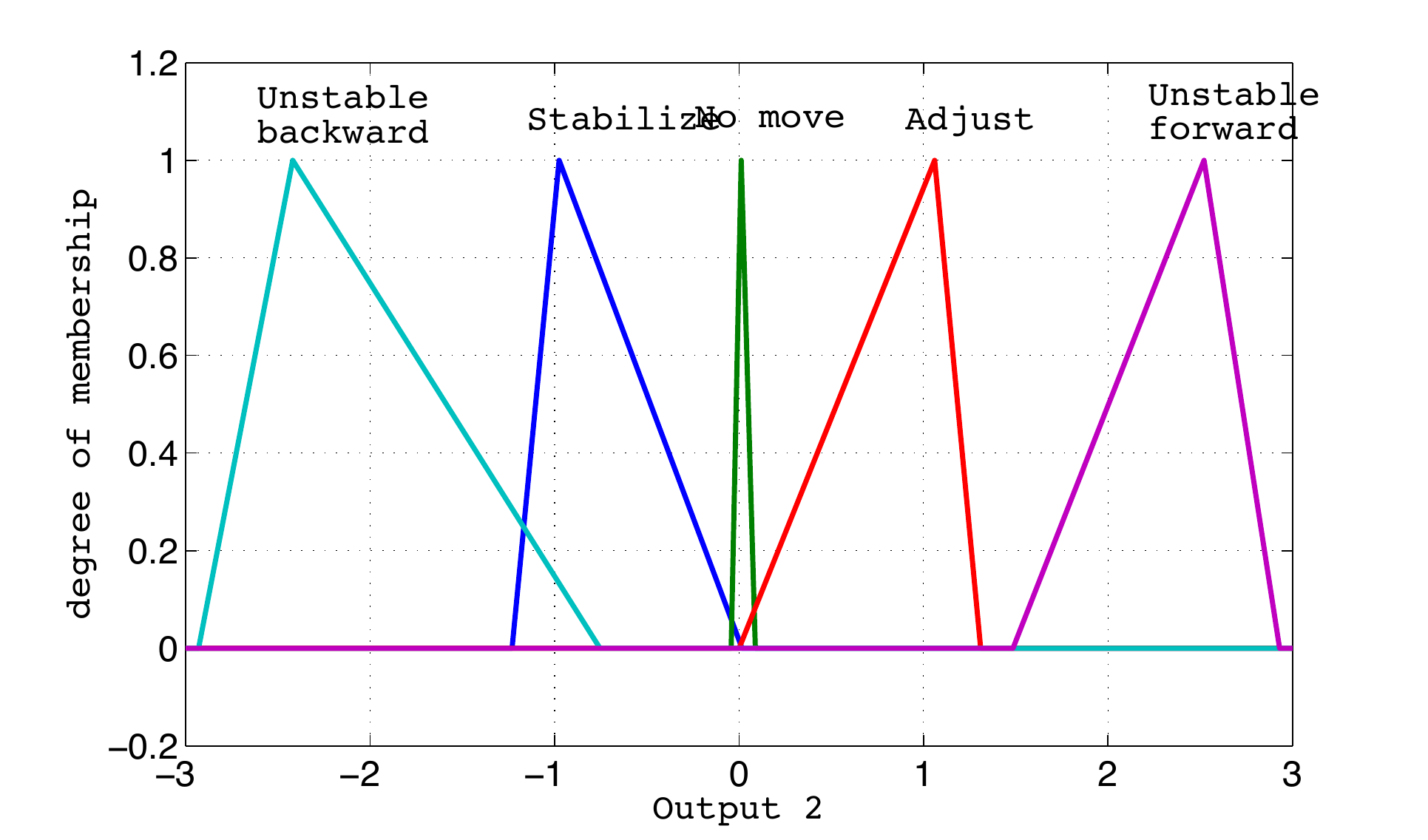}
	\caption{Membership functions for output 2}
	\label{fig:set2}
	\end{center}
\end{figure}

The membership functions for the output 2 are shown in Fig. \ref{fig:set2}, they determine the movement for
a detected phase.

The following fuzzy sets are then defined:
\begin{itemize}
	\item \textbf{Unstable forward and backward}: subject underlies high unbalance. Quick reaction is required.
	\item \textbf{Stabilize}: subject indicates desire of stabilization.
	\item \textbf{No move}: no movement is necessary in the horizontal direction.
	\item \textbf{Adjust}: subject desires another position of the handles.
\end{itemize}

If we denote high with a H, zero with a Z, low with a L, extremely low with a EL and extremely high with a EH, it is possible to explain every control mode with a fuzzy rule. As an example here is the case of RISING state: \\
\\
IF $F_{gy}$=EL AND $F_{hx}$=L AND $\frac{dF_{hy}}{dt}$=H \\
THEN the human is RISING.\\

\begin{figure}[htbp]
\begin{center}
		\includegraphics[width=8.9cm]{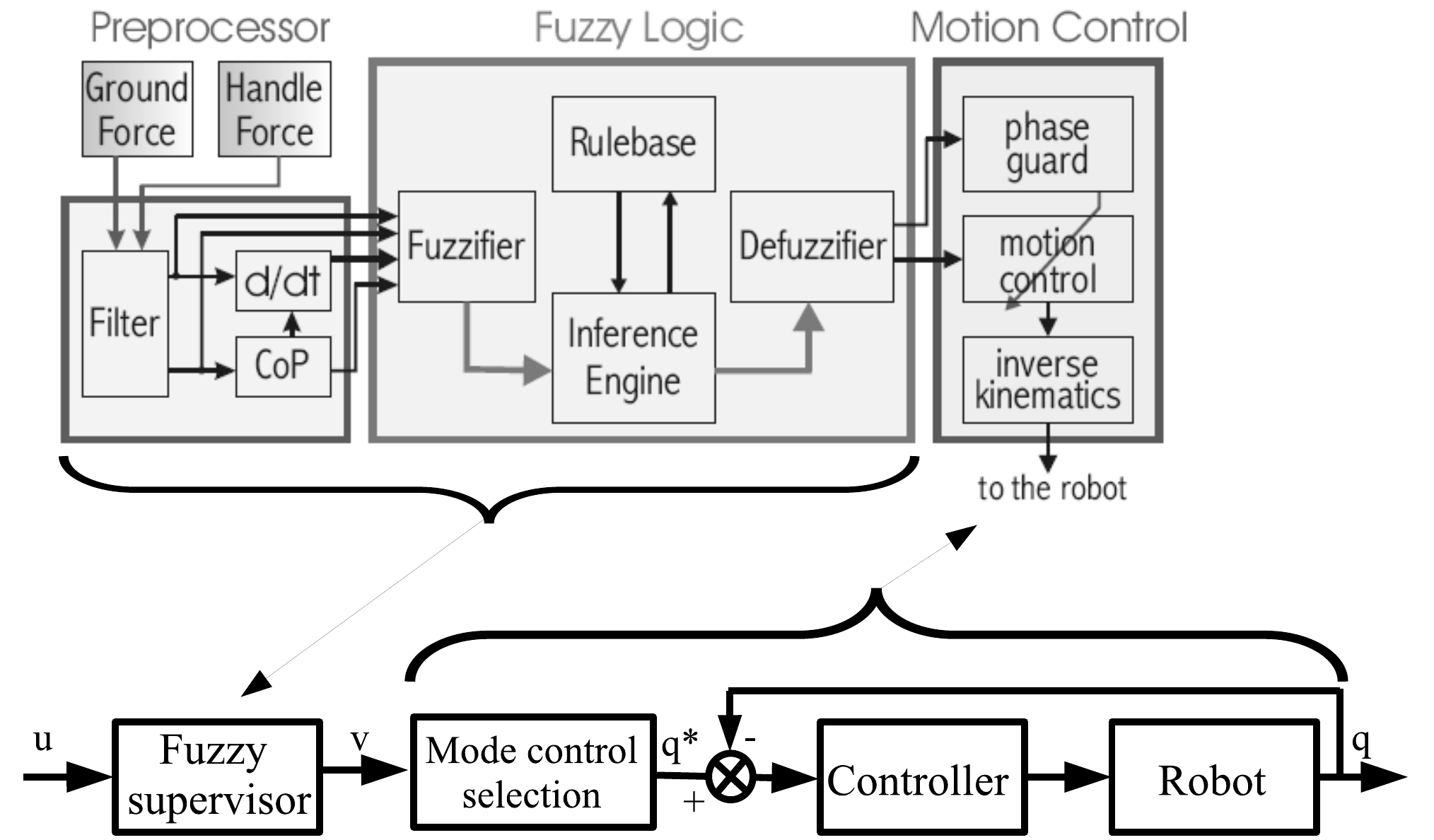}
	\end{center}
	\caption{Control structure}
	\label{fig:controler}
\end{figure}

The complete controller structure is shown in Figure \ref{fig:controler}.

Inputs of this control are Ground Forces and Handle Forces. These inputs are computed in a preprocessing block that applies a filter and calculates the position of CoP and its time derivatives. These outputs are processed by the fuzzy logic block to identify
the postural state of the patient. Then, the corresponding control mode is selected between those:
\begin{itemize}
	\item \textbf{Normal}: tracking trajectory,
	\item \textbf{Admittance}: admittance control according to handle forces measured,
	\item \textbf{Stabilization}: modification of the tracking trajectory to stabilize the patient,
	\item \textbf{Return}: the interface returns to the initial position,
\end{itemize}

\begin{figure}[!hpt]
 \centering
\subfigure{
\label{fig:FuzModel1}
 \includegraphics[width=0.21\textwidth]{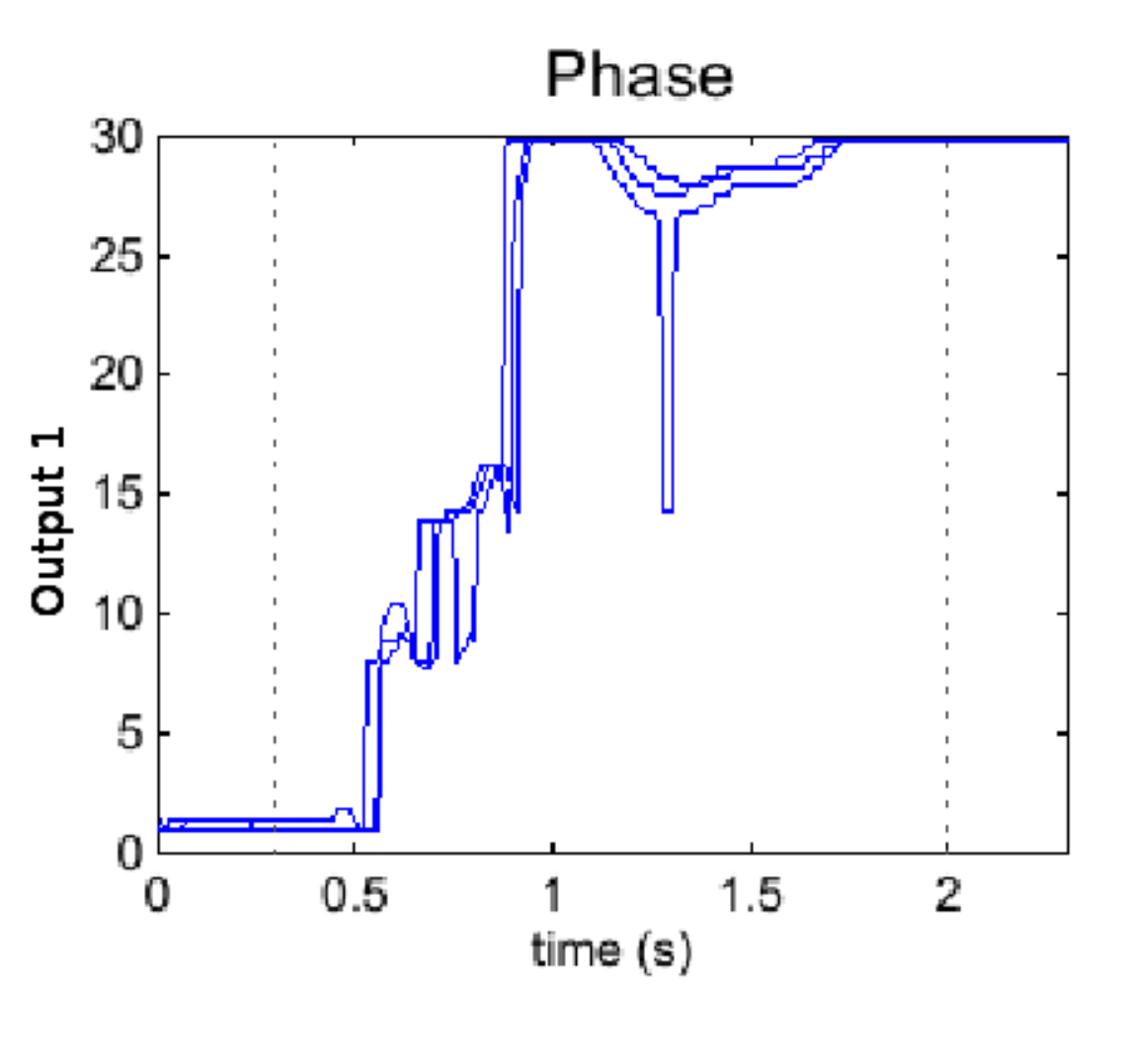}}
\subfigure{
\label{fig:FuzModel2}
 \includegraphics[width=0.21\textwidth]{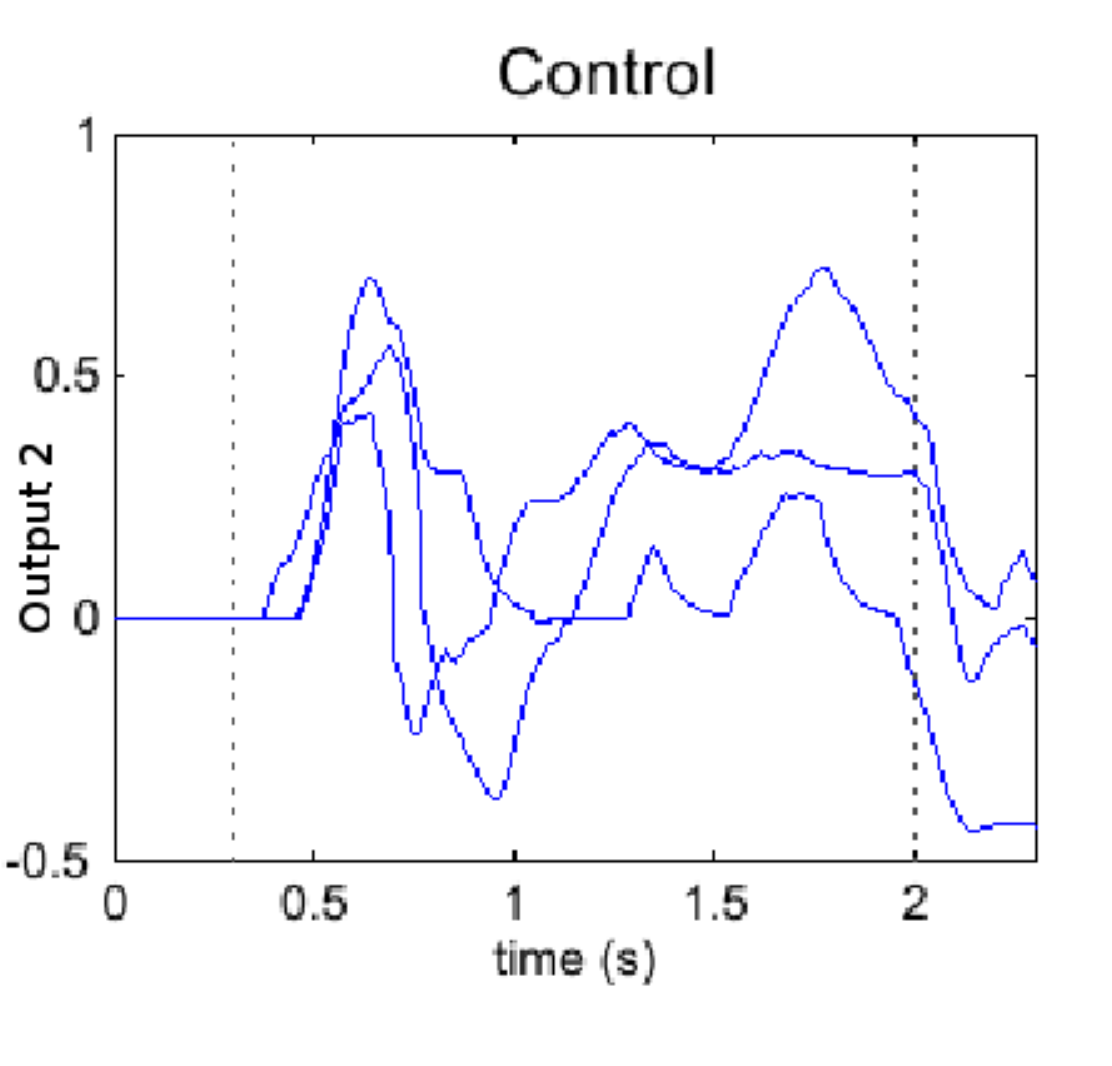}}
 \caption{Supervisor outputs: (a) represents phase identification, (b) shows stability representation}
 \label{fig:FuzModel}
\end{figure}

The outputs of this controller are represented in Fig. \ref{fig:FuzModel}. In this last figure, one can see that the supervisor can represent the different phases of the movement (Fig. \ref{fig:FuzModel1}). It can be easily  read on the picture because a normal sit-to-stand motion supposes a regular augmentation of the fuzzy output. On Fig. \ref{fig:FuzModel2} , the second output represents stability. If this output is close to zero, the postural state is stable.
That outputs have no units, their range are chosen to fit with the control needs.

\subsection{Controller}
\label{sec:controller}
From all these rules derived from analysis of sit-to-stand motion, the controller is implemented as follows.

\textbf{Admittance} control mode is a simple admittance control :
\begin{eqnarray}
\delta X= k * Fh + b * \frac{\delta X}{\Delta t\label{eq:adm}}\\
 X[t+1]=Xcur[t]+\delta X \label{eq:pos}
\end{eqnarray}
 where \textit{k} is a couple of coefficients equivalent to a spring, \textit{b} represents damp coefficients, \textit{X[t]} are the Cartesian desired position of the handles for \textit{t} time, \textit{Fh} represents Forces measured on the handles, \textit{Xcur[t]} is current coordinates of handles at \textit{t} time.

\textbf{Normal} control mode is a linear combination of admittance control and trajectory (\textit{Xtraj}) following, where \textit{output2} of the fuzzy system is a weight of admittance :
\begin{eqnarray}
\delta X=output2 * (k * Fh + b * \frac{\delta X}{\Delta t}) \label{eq:fuzadm}\\
 X[t+1]=Xtraj[t]+\delta X \label{eq:traj}
\end{eqnarray}

In the case of instability, the \textbf{stabilization control} is the admittance control, eq. (\ref{eq:fuzadm}), weighted (\textit{A}) to amplify X motions and to have no Y movement, it leads to eq. (\ref{eq:fuzadm1}). Position computation is eq. (\ref{eq:pos}).
\begin{eqnarray}
\delta X=A * output2 * (k * Fh + b * \frac{\delta X}{\Delta t}) \label{eq:fuzadm1}
\end{eqnarray}
And the trajectory is updated to fit with new situation.

The \textbf{return} control computes a linear reverse trajectory (\textit{Xrev\_traj[t]}) and comes back to initial position.
\begin{equation}
	X[t]=Xrev\_traj[t] \label{eq:revtraj}
\end{equation}

With this implemented control, the physical human-robot interaction can be evaluated on patients.

\section{Results}
\label{sec:ResDis}
Some experiments were performed on some healthy subjects to evaluate the ability of this control to help people to stand up.

\begin{figure}[hbtp]
	\subfigure{%
 		\includegraphics[width=0.12\textwidth]{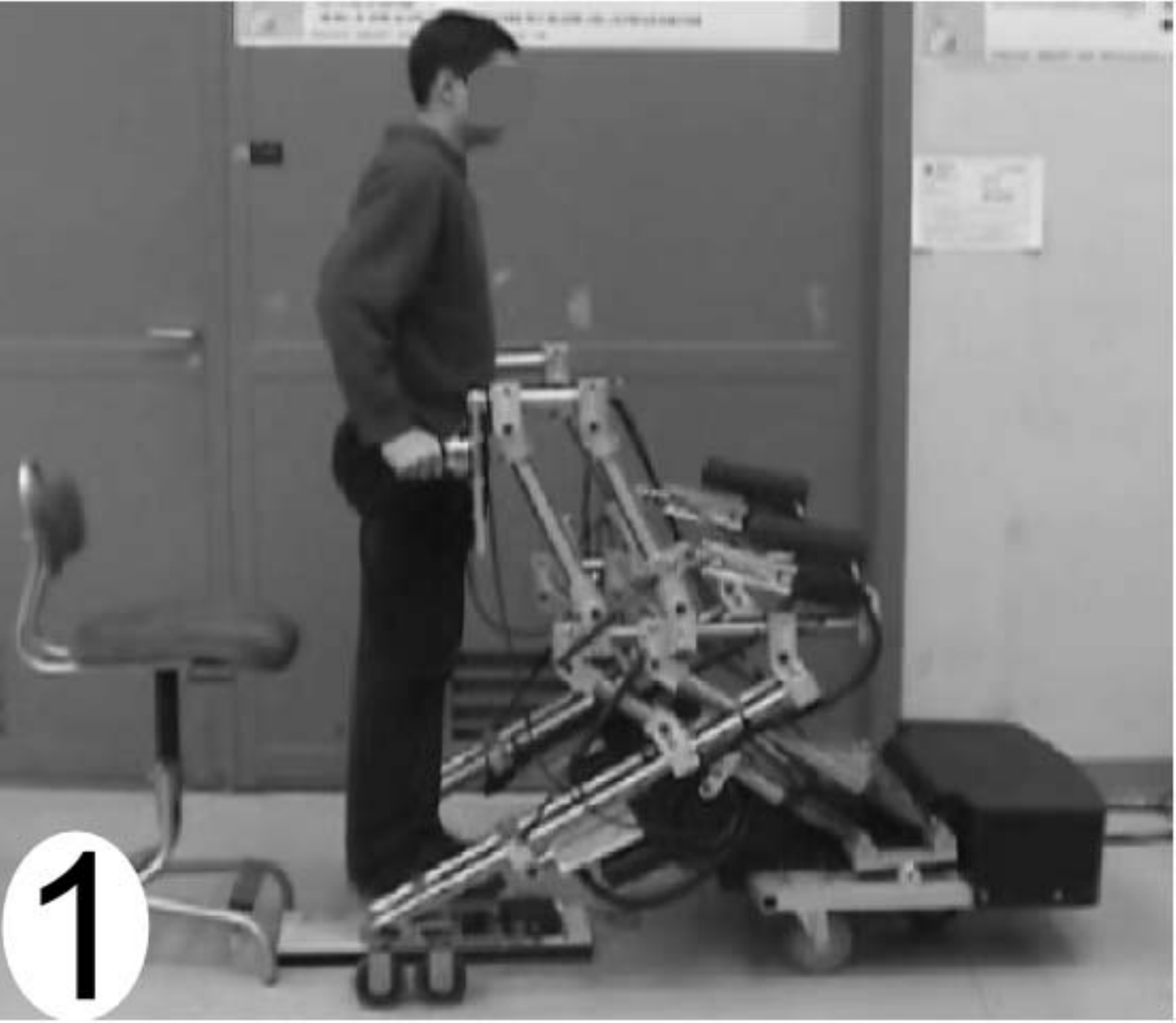}}
	\hspace{-0.01\textwidth}
	\subfigure{%
 		\includegraphics[width=0.12\textwidth]{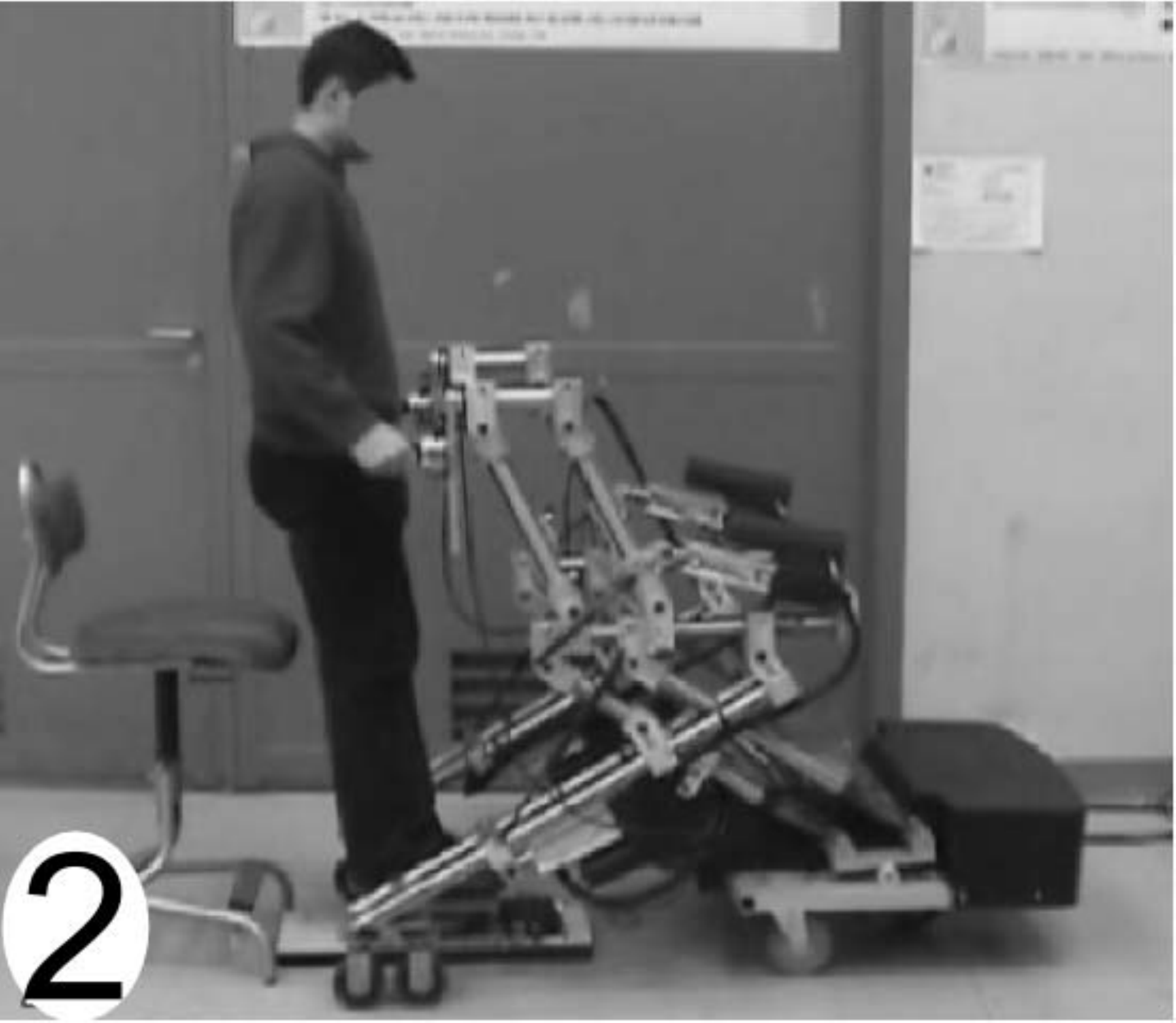}}
	\hspace{-0.01\textwidth}
	\subfigure{%
 		\includegraphics[width=0.12\textwidth]{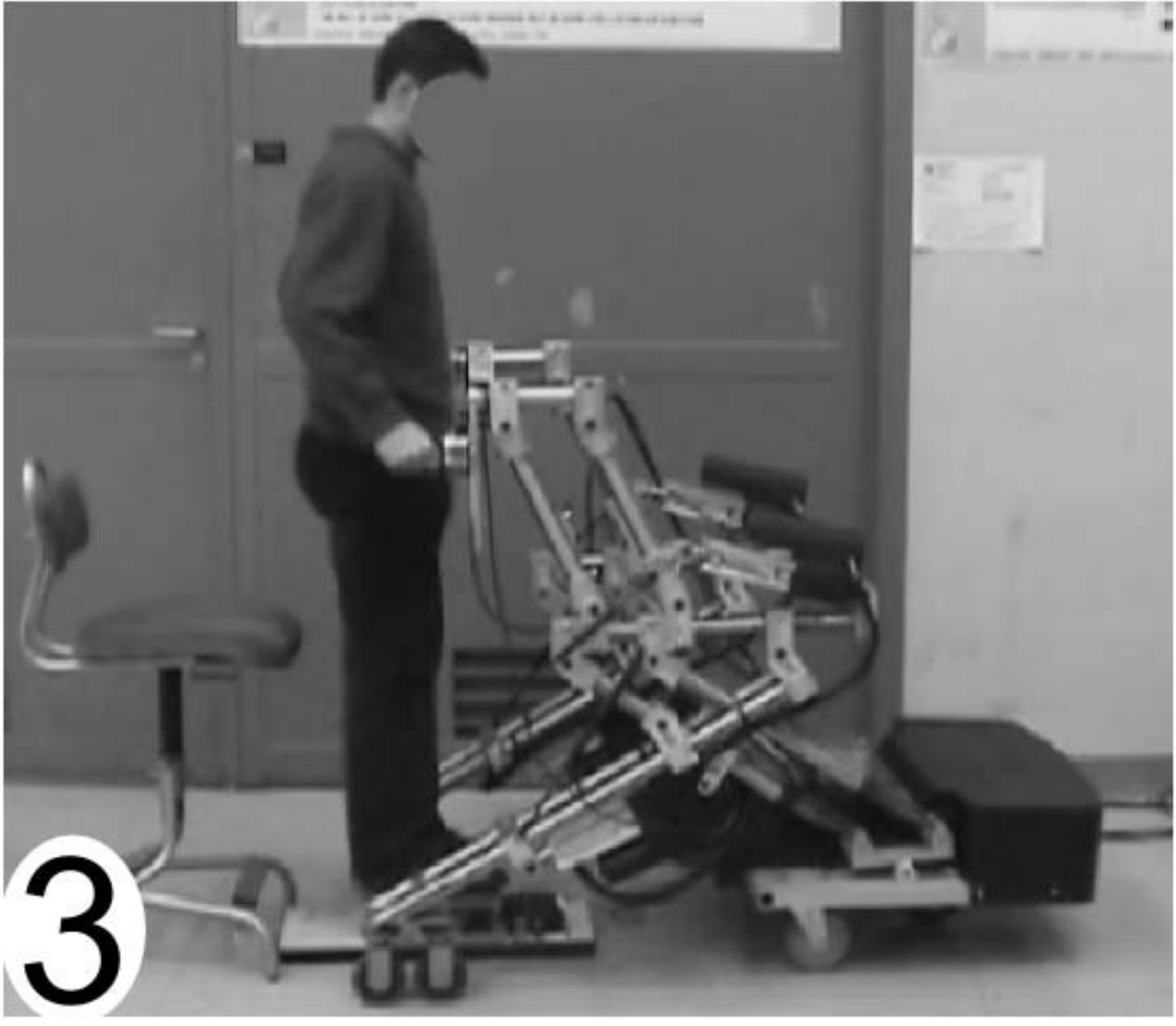}}
	\hspace{-0.01\textwidth}
	\subfigure{%
 		\includegraphics[width=0.12\textwidth]{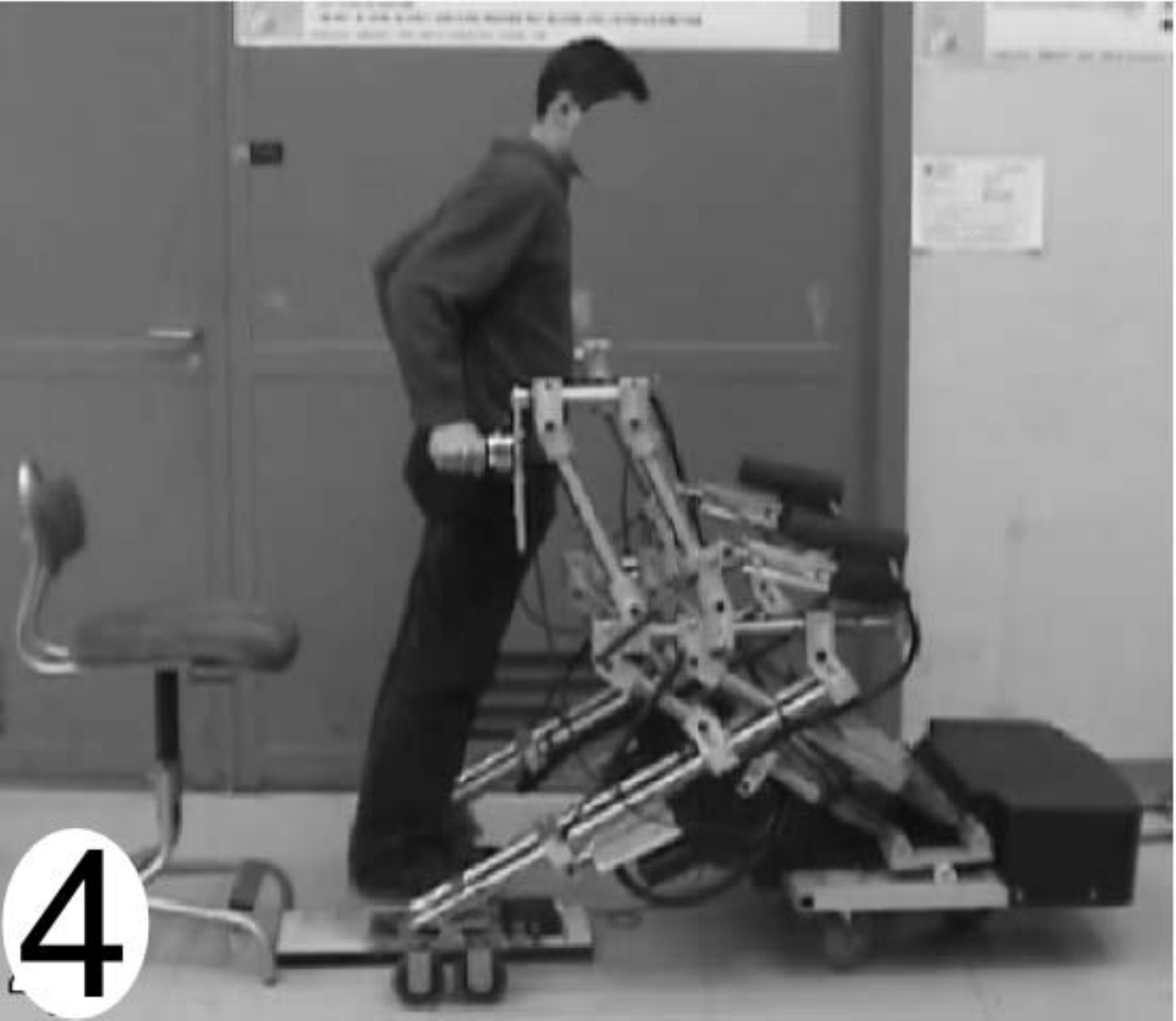}}
	\hspace{-0.01\textwidth}
	\subfigure{%
 		\includegraphics[width=0.12\textwidth]{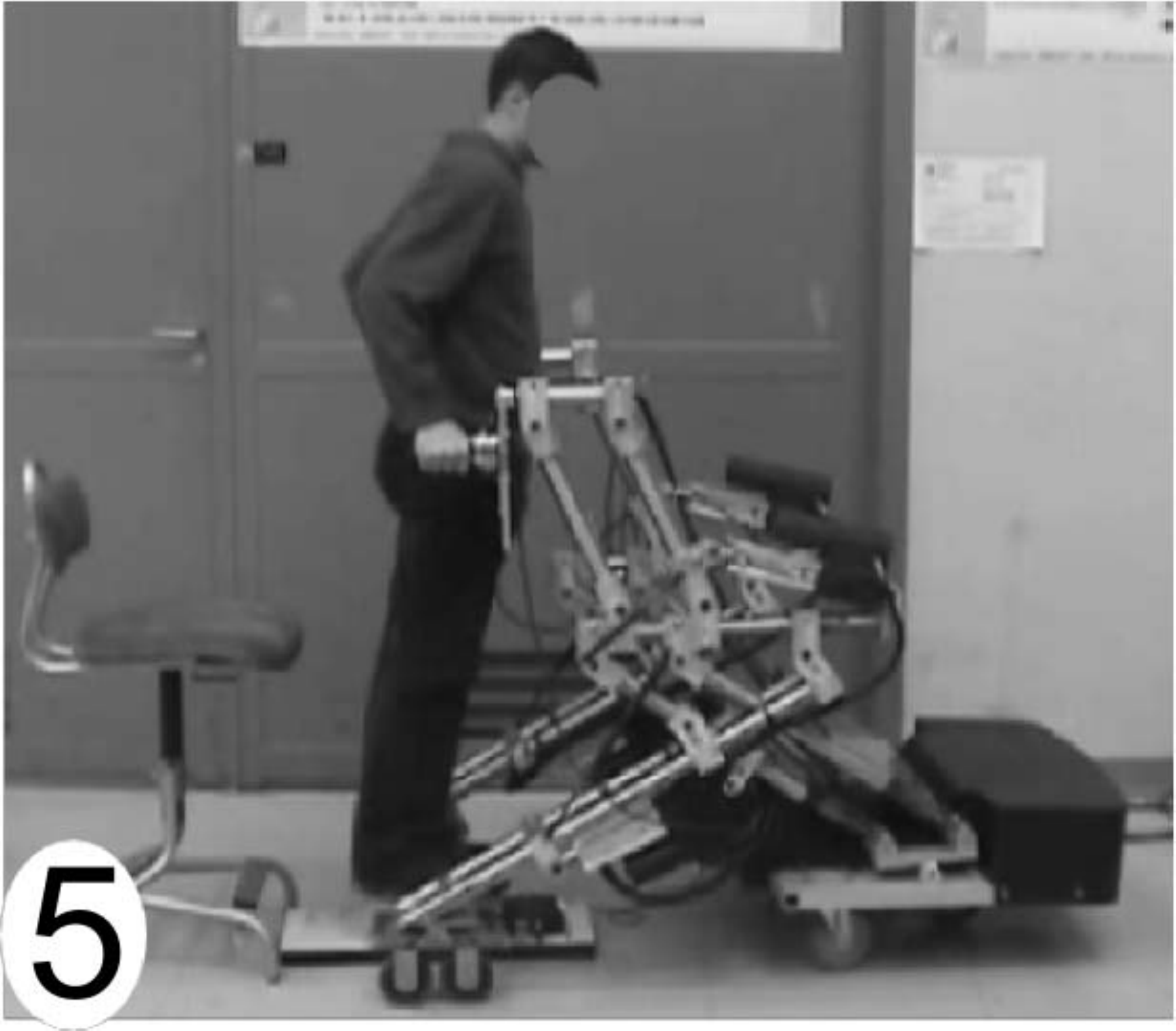}}
	\hspace{-0.01\textwidth}
	\subfigure{%
 		\includegraphics[width=0.12\textwidth]{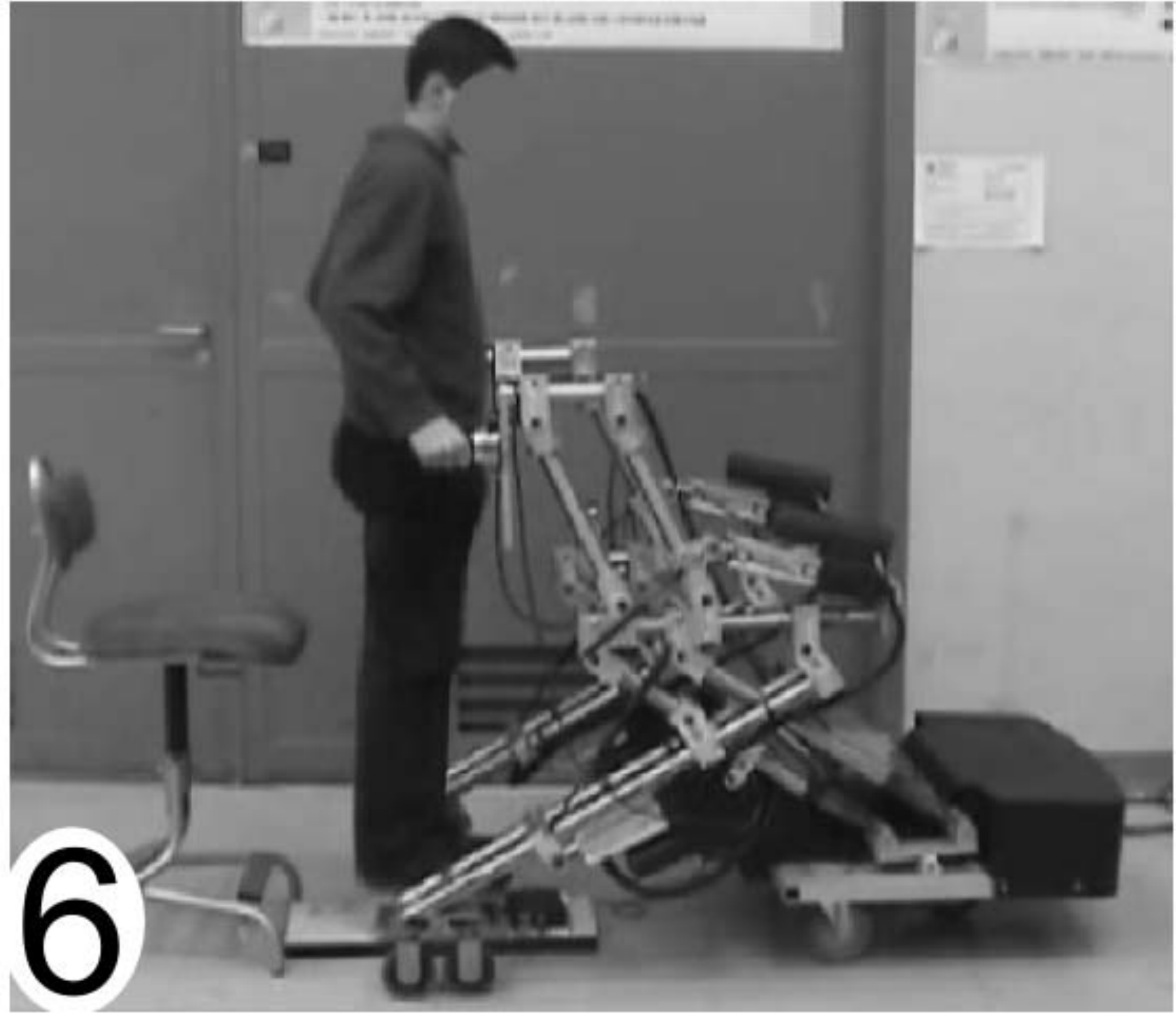}}
	\hspace{-0.01\textwidth}
	\subfigure{%
 		\includegraphics[width=0.12\textwidth]{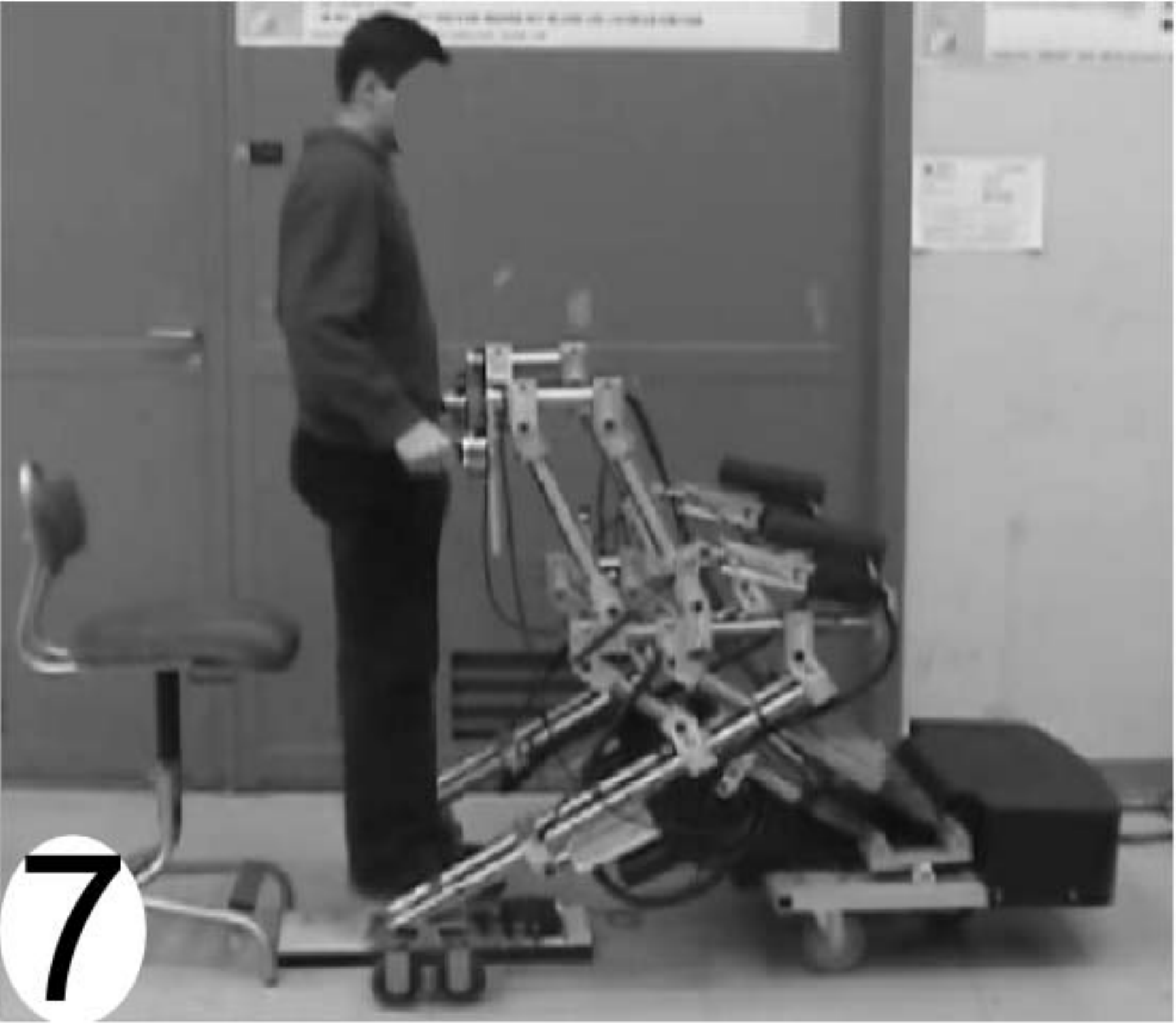}}
	\hspace{-0.01\textwidth}
	\subfigure{%
 		\includegraphics[width=0.12\textwidth]{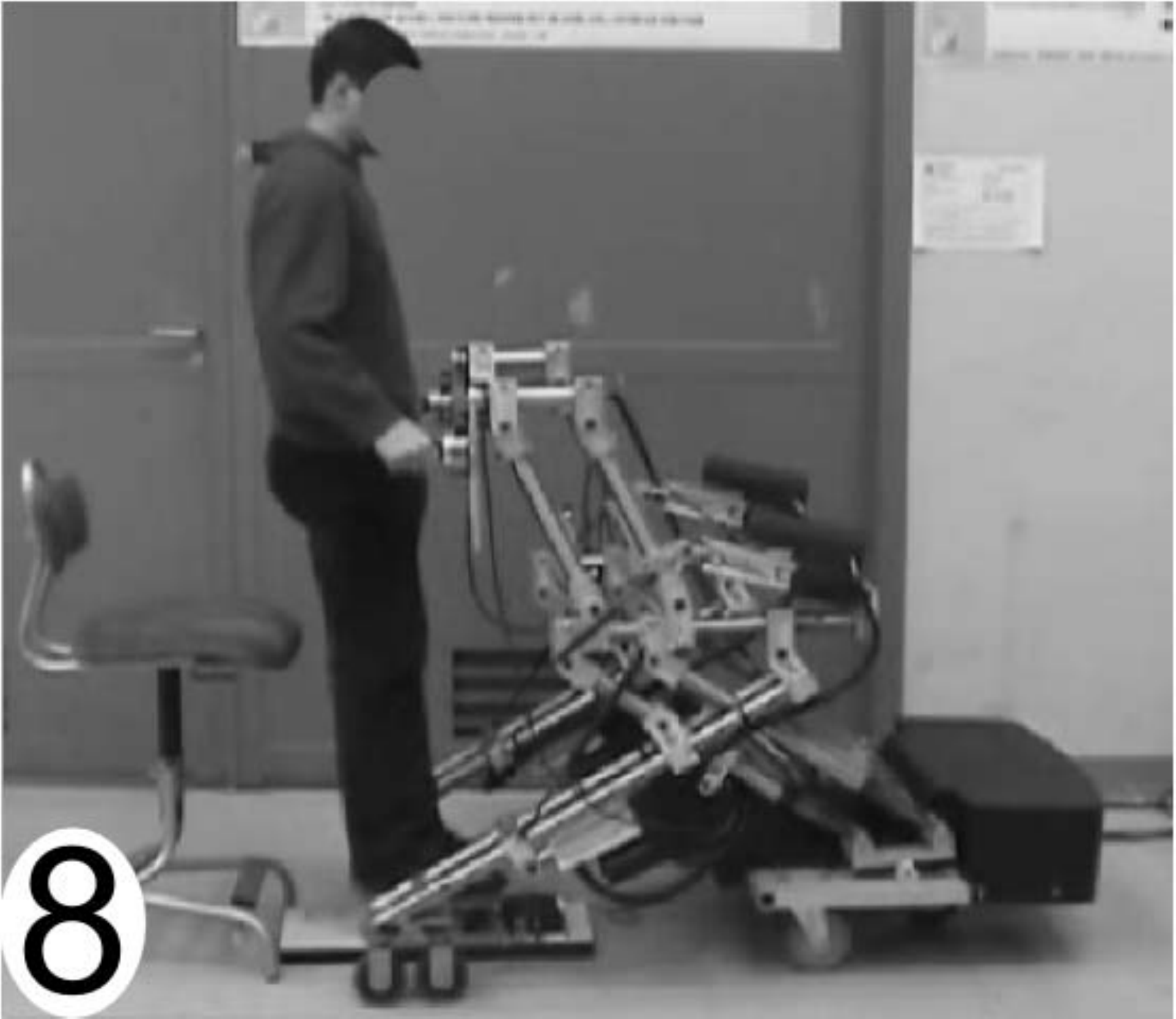}}
	\hspace{-0.01\textwidth}
\caption{Robotized support for instability control}
\label{fig:realInsta}
\end{figure}

Laboratory experiments were performed to evaluate action of stability control (fig.\ref{fig:realInsta}).
Those experiments on instability were not performed on patients, because there is not an acceptable scenario that can safely put the light on the instability control with patients.

\subsection{Ethical Issue}
From a legal point of view, trials presented in this paper are not medical trials but technological trials.
However, some consequences are still remaining furthermore when one works with patients. So, these experiments are bound to have an ethical approach, that includes having an external overview of the whole testing process, in our case it is done by medical staff.
According to their doctors, all subjects are able to consent to participate and their written agreement are asked.
All data are anonymous for processing, any personal data is stored in a secured location different from the anonymous data location.

\subsection{Clinical Results}

The robotic device with its control was evaluated on patients in 'URF-Bellan'.
The patients are affected by multiple sclerosis. In many cases, multiple sclerosis patients present cerebellar ataxia, that affects their motion with some tremors that can lead to disbalance and fall.
The criteria include patients that have:
\begin{itemize}
	\item static or kinetic cerebellar syndrome of lower limbs
	\item no or minor force deficit in upper limbs
	\item sufficient muscular force in lower limbs allowing the every day life sit-to-stand motion with human help
\end{itemize}

The fuzzy controller was evaluated on 10 patients presented in table \ref{tab:DataDesc}.\\
This group of patients is composed of 6 males and 4 females. The average age of males (resp. females) is 36.5 years (resp. 51 years) with a standard deviation of $\pm$7.12 years (resp. $\pm$18 years). The average weight of males (resp. females) is 78.8 kg (resp. 59.7 kg) and the mean size of males is 1.80 m.
\begin{table}[!htb]
  \centering
\begin{tabular}{|c|c|c|c|c|}
\hline
ID & Gender & height & weight & age \\
\hline
Patient1 & M & 1.90 & 85& 40\\
Patient 2  & M & 1.75& 74& 39\\
Patient 3  & M & 1.77 &71& 24 \\
Patient 4  & M & 1.84 &86& 37 \\
Patient 5  & M & 1.89& 94& 34 \\
Patient 6  & M & 1.65 &63& 45 \\
\hline
Patient 7  & F & 1.60 && 60? \\
Patient 8  & F &  1.59&54& 24\\
Patient 9  & F & & 60 & 60? \\
Patient 10 & F & & 65 & 60?\\
\hline
\end{tabular}
\caption{Patients description}
\label{tab:DataDesc}
\end{table}

The synthetic table (\ref{tab:ResDesc}) shows the achievement of task (sit-to-stand) by patients supported by the robotic device controlled by our fuzzy logic controller.
\begin{table}[!htb]
  \begin{center}
\begin{tabular}{|c|c|c|}
\hline
{\tiny ID}  & {\tiny \# of trials} & {\tiny \# of achieved  }\\
\hline

Patient1  & 10 &8 \\
Patient 2& 15& 14  \\
Patient 3 &10 & 9  \\
Patient 4& 11 &9  \\
Patient 5 & 12 &11  \\
Patient 6 & 13 &11 \\
\hline
Patient 7* & 2& 1*  \\
Patient 8& 12& 10 \\
Patient 9   &5 & 4 \\
Patient 10 & 7& 6 \\
\hline
\end{tabular}
\caption{Patients data and assisted sit-to-stand achievement}
\label{tab:ResDesc}
\end{center}
\textit{\footnotesize{*the case of patient 7 is particular because she was affected of spasticity of one leg, feet sensor were unusable so fuzzy based control was not possible.}}
\end{table}

\begin{figure*}
 \centering
\subfigure[ Filtered output $\nu_1$ (Bold Line) Versus Direct Output $\nu_1$ of the supervisor]{
 \label{fig:FuzRes1_Filt}
 \includegraphics[width=0.32\textwidth]{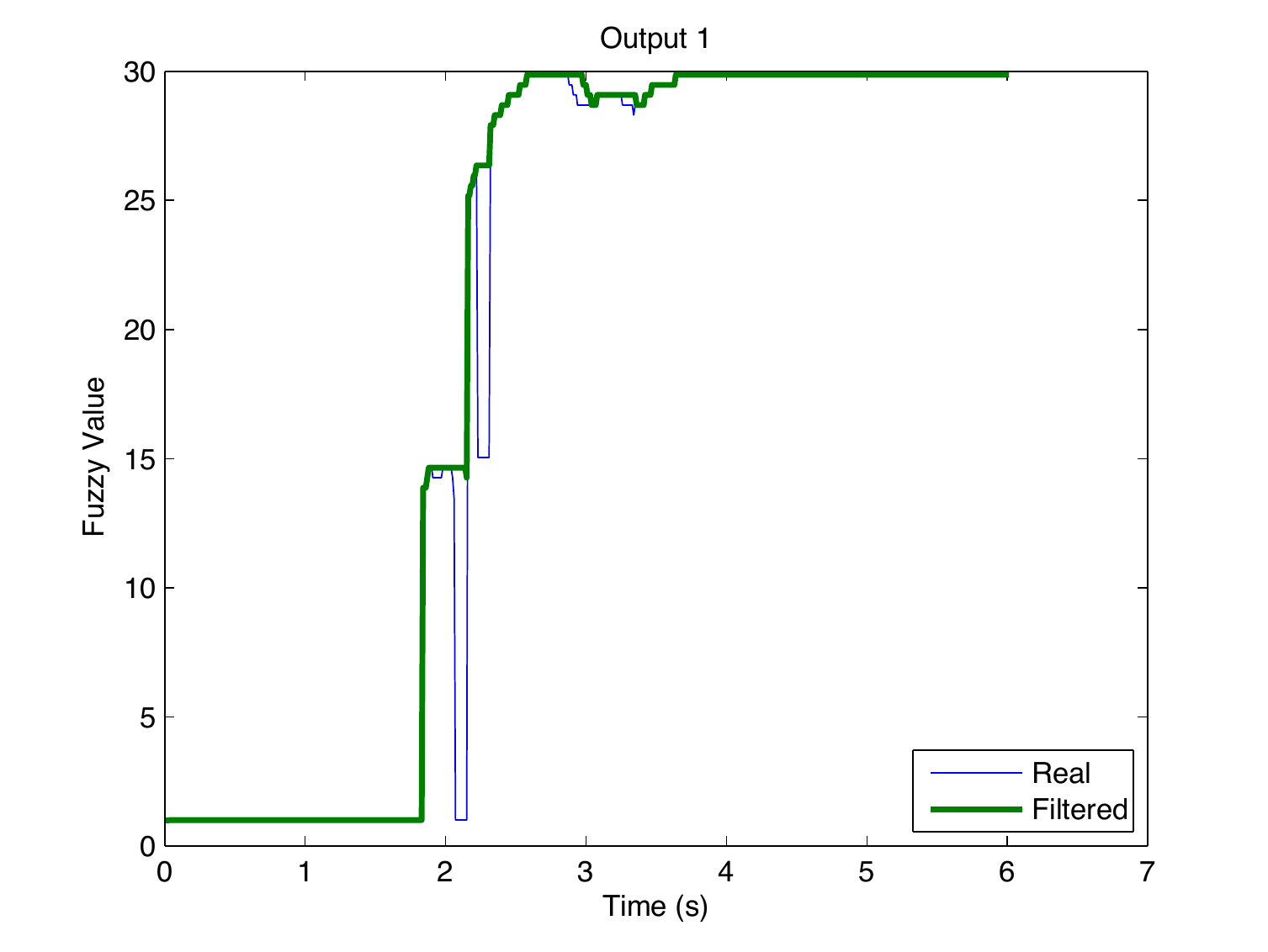}}
\subfigure[Supervisor output 2 during a sit-to-stand motion of patient 8]{
 \label{fig:FuzRes2}
 \includegraphics[width=0.32\textwidth]{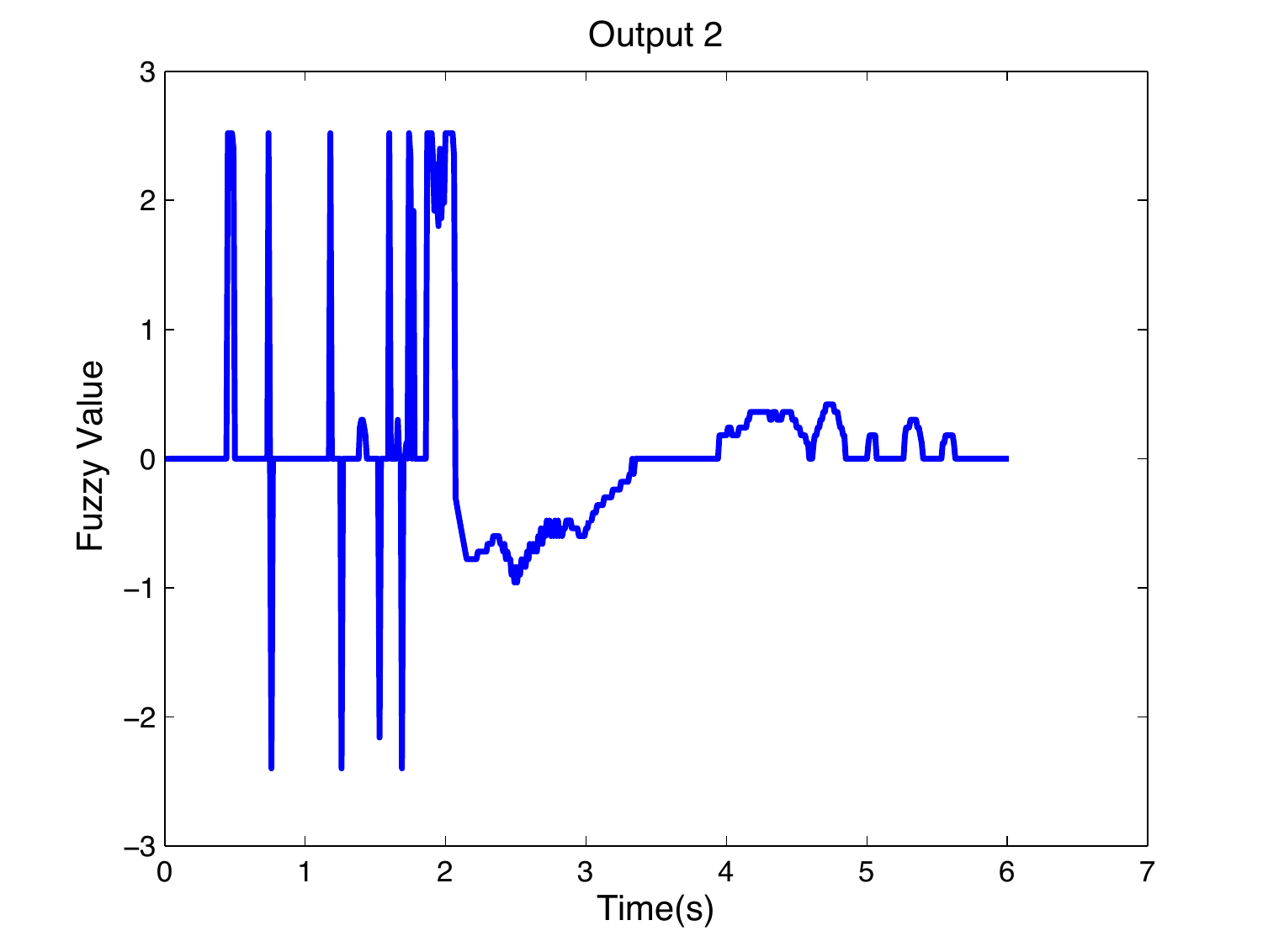}}
\subfigure[Supervisor output 1 during a sit-to-stand motion of patient 8]{
\label{fig:FuzFor1}
 \includegraphics[width=0.32\textwidth]{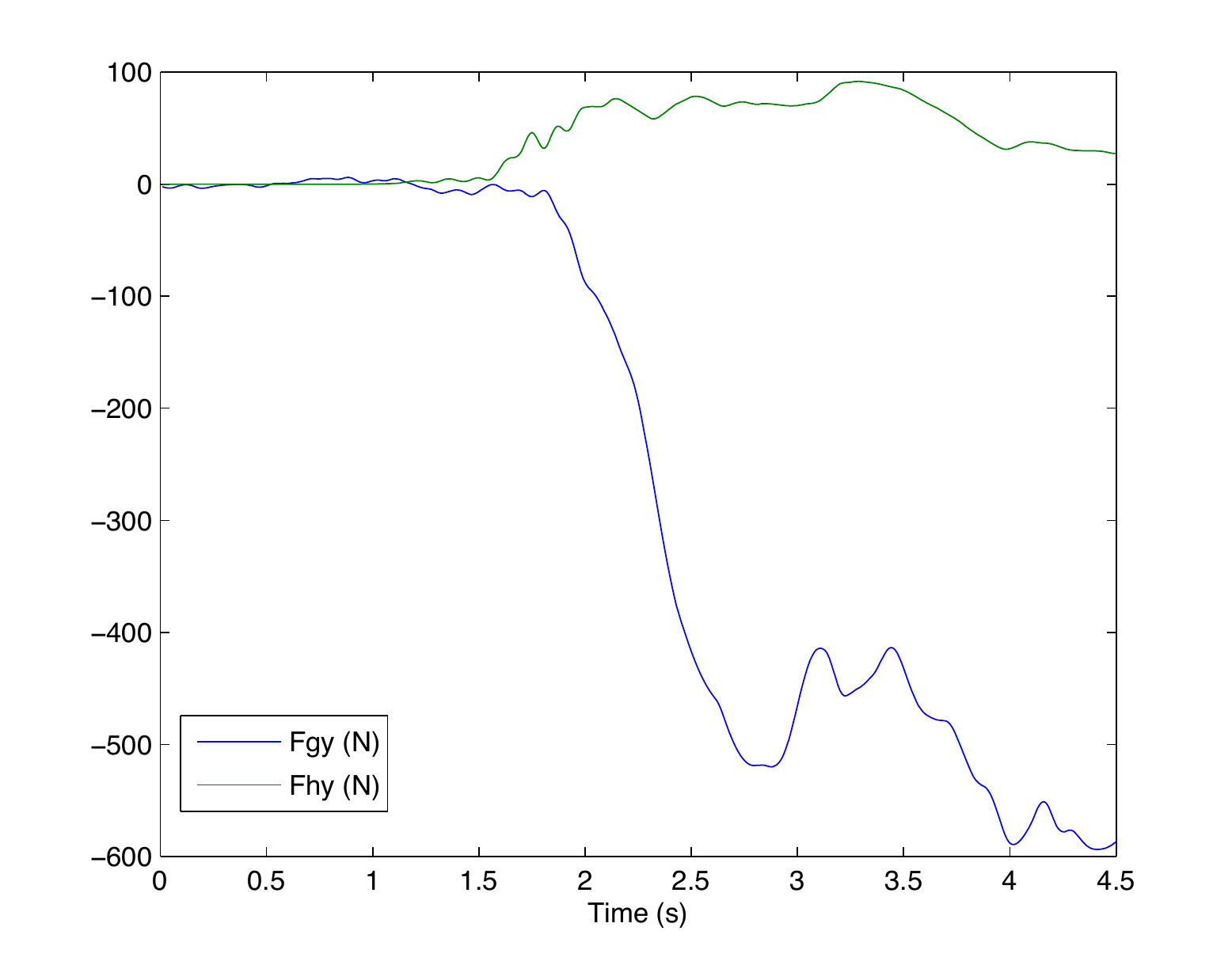}}
\caption{Outputs of different solutions}
\end{figure*}

As we described in the previous part, two fuzzy outputs are managed by the supervisor.
The first output represents different phases of the sit-to-stand movement and is presented in Fig. \ref{fig:FuzRes1_Filt} line called ``real''.
This picture shows a sit-to-stand motion achieved by the patient without any difficulty or hesitation. However, we can see on this picture that phases of sit-to-stand motion are not well described by the supervisor, indeed the real outputs of the supervisor present some discontinuities in comparison to the motion of the simulated supervision (Fig. \ref{fig:FuzModel}).

To fix that, we improved the control with a filter on the supervisor output checking if the output $\nu_1$ used in the control is the maximum value of the ten preceding values of the output $\nu_1$ of the supervisor. This filter gives in the worst case a delay of 100 ms in the control but this delay does not impact the good use of the robotic device. The result of this filtering is presented in Fig.~\ref{fig:FuzRes1_Filt} and is compared to the output result of the supervisor.

The second output (Fig.~\ref{fig:FuzRes2}) represents stability of the patient during the motion. As one can see, a lot of noise appears when the patient is sitted. This noise is not a problem because the supervisor is switched to a control for sitted patient that does not use stabilisation information. Indeed, when a patient is sitted, the controller must choose between null-effort control for repositioning handles or initialisation of rising motion.
At the beginning of the sit-to-stand movement (near time 2s), one can notice that instability is high and output1 stays near 15. In this situation the controller is in \textbf{instability control mode} and as it is explained in section \ref{sec:controller} the trajectory is stopped and there is an admittance control along the $\vec{x}$ direction. Then the trajectory is updated, as far as this is in the beginning, the trajectory is only shifted along the $\vec{x}$ direction..

\begin{figure*}[htbp]
 \centering
\subfigure[]{%
\label{fig:sts1}
\includegraphics[width=0.15\textwidth]{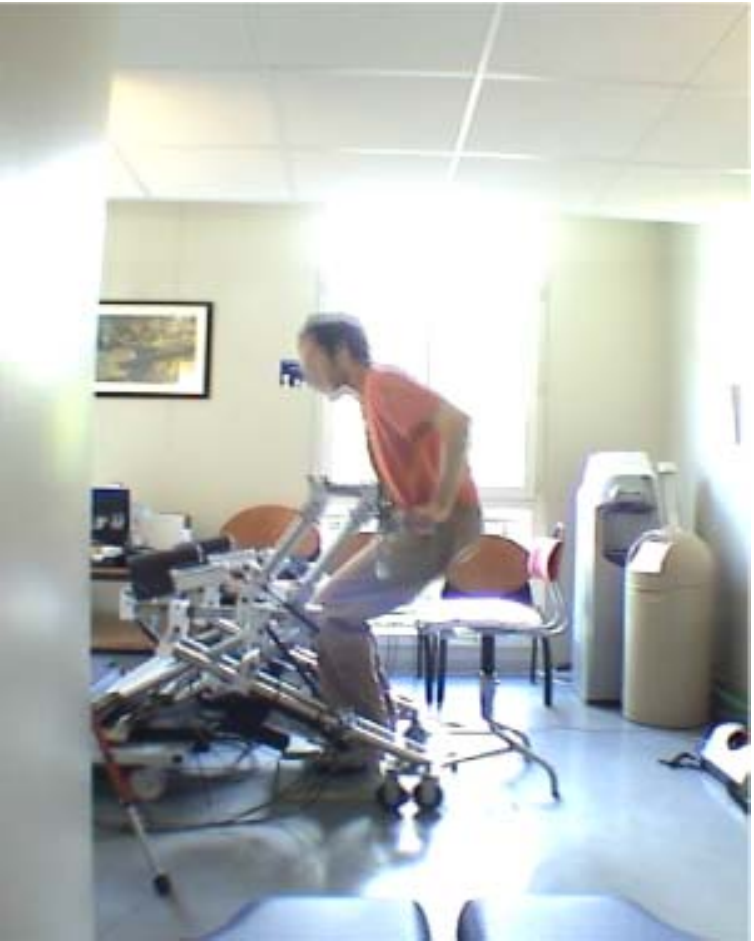}}
\subfigure[]{%
\label{fig:sts2}
\includegraphics[width=0.15\textwidth]{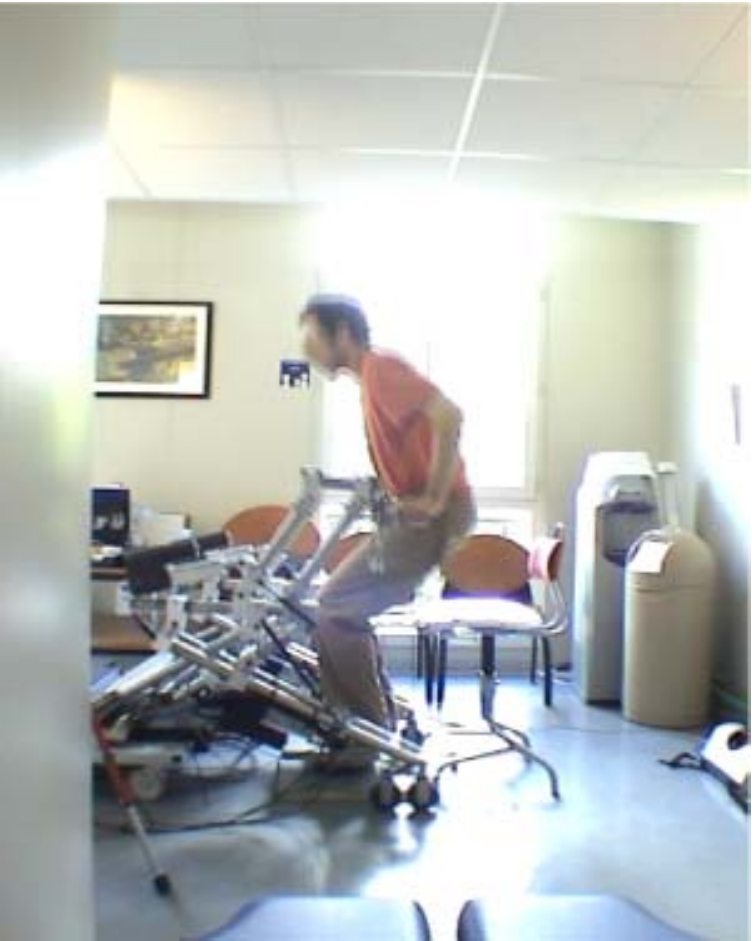}}
\subfigure[]{%
\label{fig:sts3}
\includegraphics[width=0.15\textwidth]{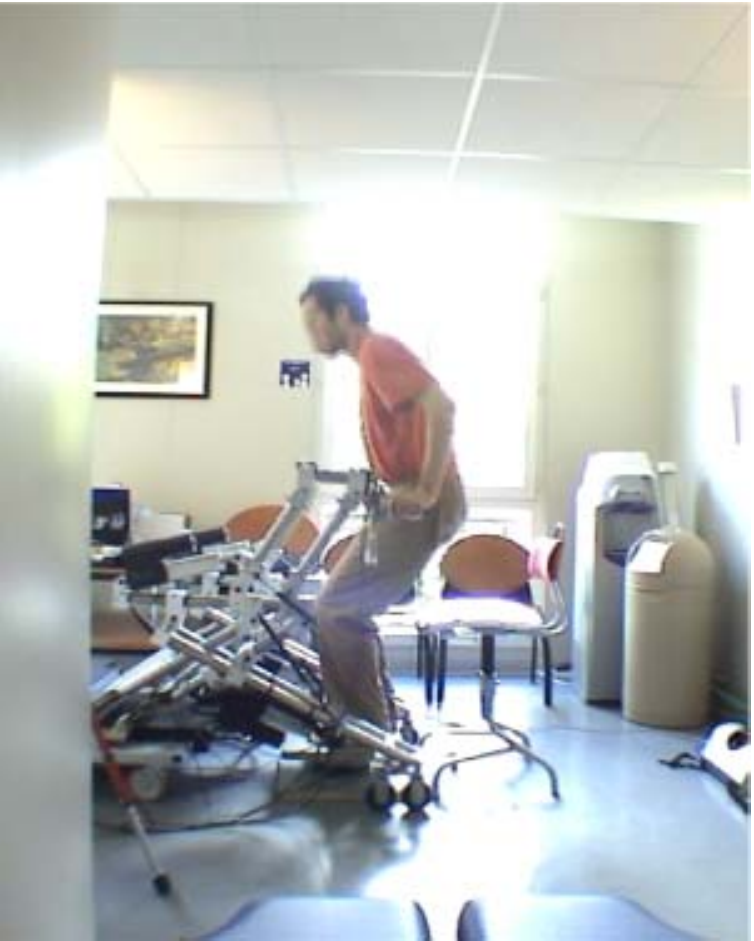}}
\subfigure[]{%
\label{fig:sts4}
\includegraphics[width=0.15\textwidth]{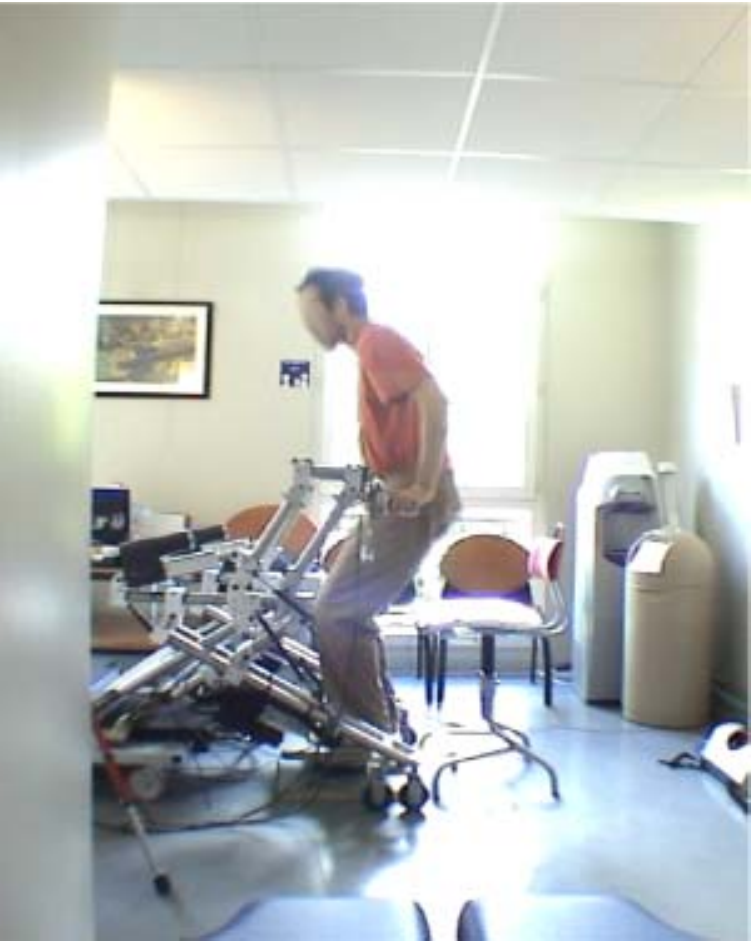}}
\subfigure[]{%
\label{fig:sts5}
\includegraphics[width=0.15\textwidth]{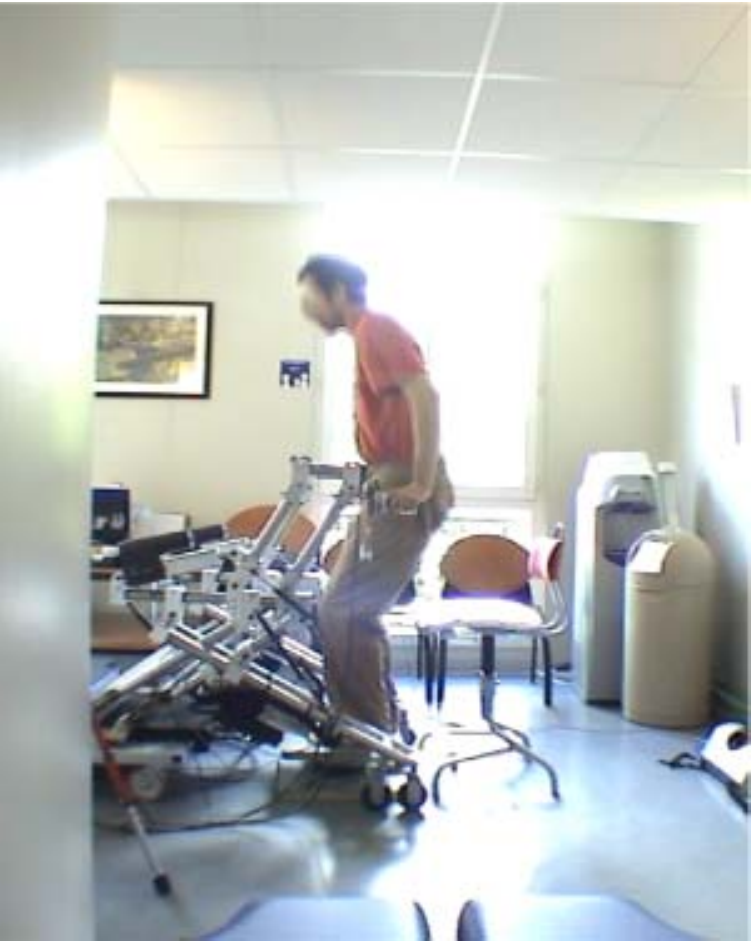}}
\subfigure[]{%
\label{fig:sts6}
\includegraphics[width=0.15\textwidth]{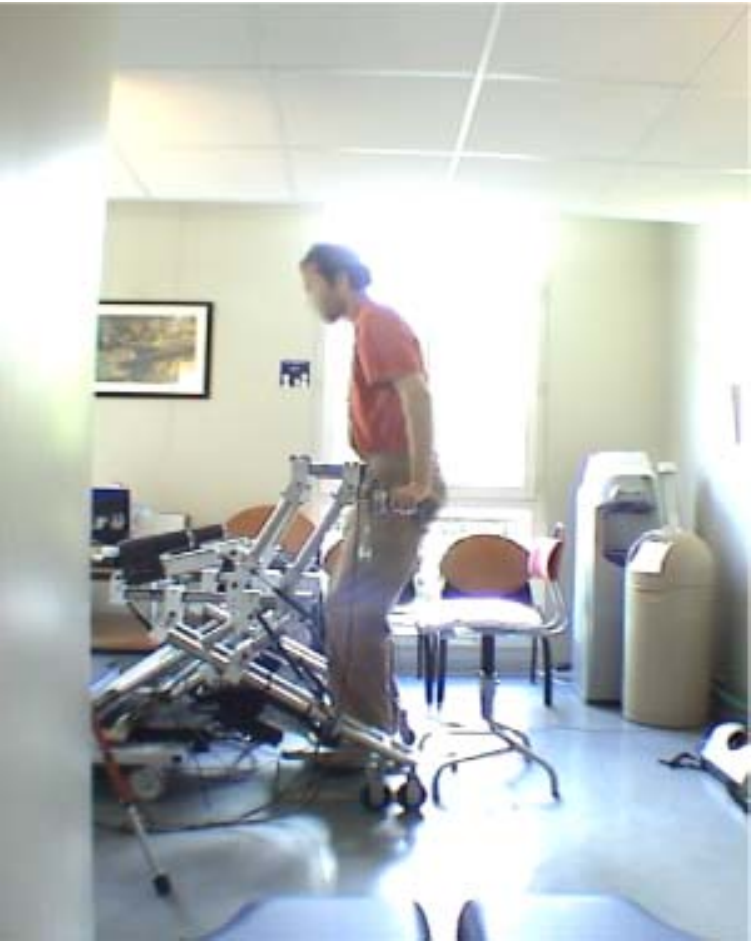}}
\caption{Patient 3 automatically supported during his sit-to-stand transfer }
\label{fig:expMovie}
\end{figure*}

As we can see in Fig. \ref{fig:expMovie}, the controller is able to help a patient to stand-up. Note that the wheels are moving during the sit-to-stand motion in order to reduce the sustentation of the robot.
It is also important to recall that all this motion is automatically controlled by the action of the patient.

\begin{figure*}[htbp]
 \centering
 \includegraphics[width=0.8\textwidth]{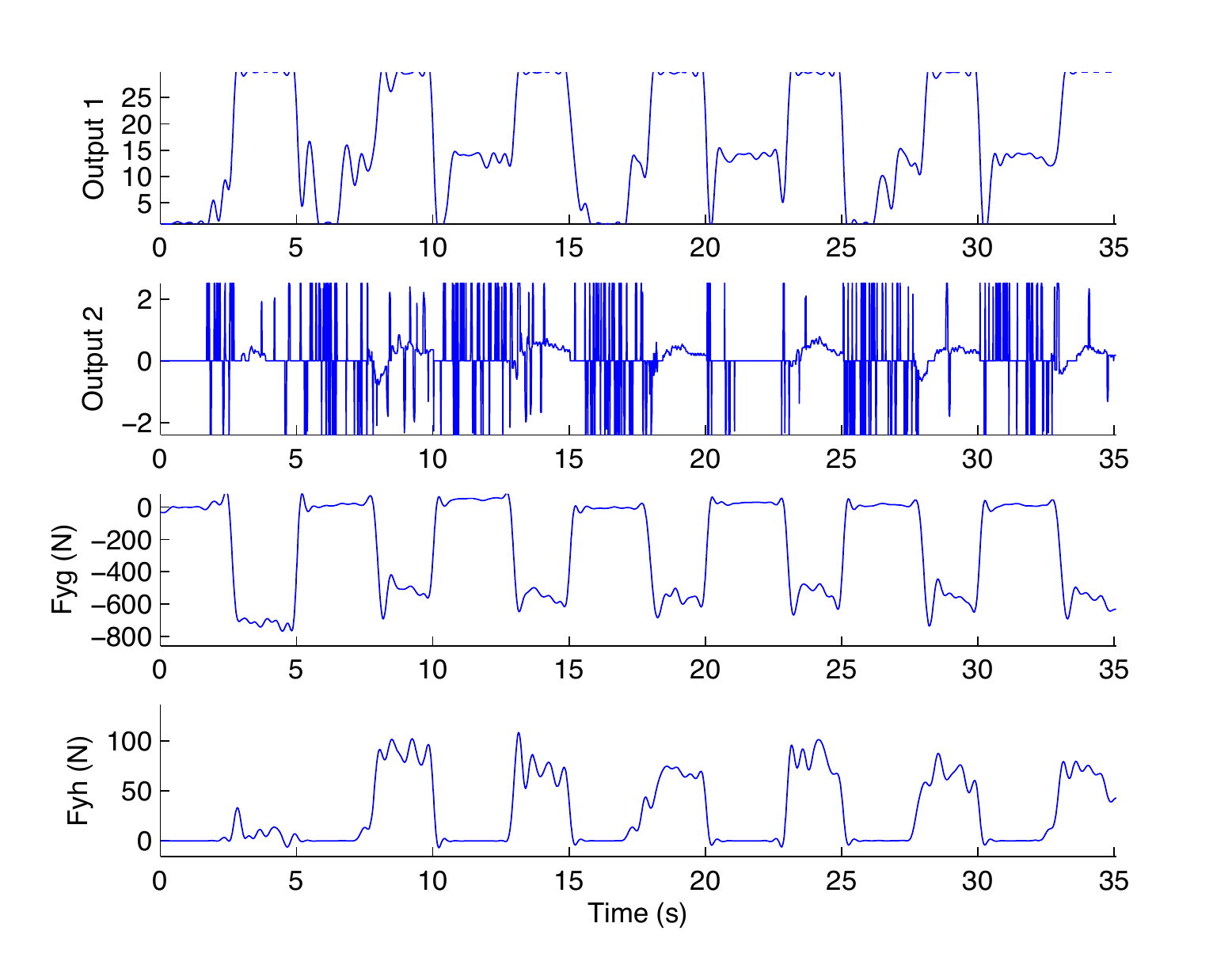}
 \caption{Records of a complete experimental set with patient 1}
 \label{fig:AllVertis}
\end{figure*}

A whole experiment done on patient 1 is presented in Fig.~\ref{fig:AllVertis} and one sample of sit-to-stand y-forces is presented in Fig.~ \ref{fig:FuzFor1}.
As one can see on these figures, forces used on the handle (Fhy) are lower than 100 Newtons and the overall maximum value obtained is 260 Newtons. The shape of these forces seems similar to those recorded during human-human sit-to-stand experiments (Fig. \ref{fig:stsana}).

During these clinical experimentation, patients learn very quickly to use the robotic device. Indeed the average number of failures while using the robot in the sit-to-stand protocol is around 1, the reason is often due to a bad positioning of the handles at the beginning. The natural position chosen by the patient is firstly too high and too far away from the trunk. That position is consequently too hard to maintain because it needs too much strength in the hands to support weight.
 After a failure, patients are advised to position the handles of the robot near the sides of the hips.  And, when this position is used, the patient is able to stand-up without any trouble.

\section{Discussion}
\label{sec:discuss}
Human-robot interaction implies an increase of noise coming from electrical power, another origin of noise is action of the robot. When the robotic device moves, dynamical properties are far from movement coming from a human in action. However, as the method is robust, as it is able to work through these difficulties.

As it is explained in part \ref{sec:ResDis}, these noises leaded to some improvement in the control implementation (filter on output 1). These improvements bring some delays in the state evaluation but it has been shown that it is completely accepted by the patients. 

The good learning rate is an interesting property of the system that is in our view due to the human centered design. All the design is done with the main idea to support a human body. The choice of handle is preferred to be more natural. Sensors chosen do not need any wearing of equipment. Above all, the control is guided by motions which are as natural as possible and with the simplest possible communication. These choices lead to a robotic device that implies a very small cognitive load for the patient that helps patients to focus on their movement rather than on the device. 

However all these results are based on 10 patients and for 10 motions. It is a first step that can lead to the conclusion that this device seems able to support patients but this protocol needs to be experimented in hospital during years to really assess the rehabilitation ability of our system and also to bring proof of the ability of this system.

Another limitation of this work is the way fuzzy parameters are tuned. Indeed, tuning is based on a small set of data coming from healthy subjects. There is room in this part for optimization on the way these parameters are tuned. In the same order of idea, it can also be interesting to propose some optimisation strategies for the whole control tuning.

Finally the use of a ground force sensor becomes a limitation when we imagine protocols that combine sit-to-stand motions with walking. We need to develop some solutions that are able to work without a force sensor under the feet.

\section{Conclusion}
It has been shown in this paper that our rehabilitation robotics device with its fuzzy based control seems able to assist patients in sit-to-stand motions. 
The fuzzy-based control benefits from using a supervisor in the control to identify states of the human motion and determine the corresponding best strategy. This kind of control results in a natural style of interaction where each partner interacts physically with the other and a common movement emerges from this interaction, that improves the feeling of the patient. This reactive and interaction based kind of control is a promising approach to rehabilitation.

\bibliographystyle{IEEEtran}
\bibliography{./Biblio.bib}
\end{document}